\documentclass[11pt]{article}

\usepackage[preprint]{acl}

\usepackage{times}
\usepackage{latexsym}

\usepackage[T1]{fontenc}

\usepackage[utf8]{inputenc}

\usepackage{microtype}
\usepackage{inconsolata}
\usepackage{graphicx}
\usepackage{amsmath}
\usepackage{amssymb}
\usepackage{booktabs}
\usepackage{array}
\usepackage{algorithm}
\usepackage{algpseudocode}
\usepackage{placeins}

\title{SEAGym: An Evaluation Environment for \underline{S}elf-\underline{E}volving LLM \underline{A}gents}

\author{
{\normalsize\bfseries
Congjie Zheng\textsuperscript{1,*} \quad
Chuanyi Xue\textsuperscript{1,*} \quad
Bin Liang\textsuperscript{1} \quad
Jun Yang\textsuperscript{1,\textdagger} \quad
Changshui Zhang\textsuperscript{1,2,\textdagger}
} \\
[-0.15em]
{\small\normalfont
\textsuperscript{1}Department of Automation, Tsinghua University, Beijing, 100084, China
} \\
[-0.15em]
{\small\normalfont
\textsuperscript{2}Beijing National Research Center for Information Science and Technology (BNRist),
} \\
[-0.15em]
{\small\normalfont
Tsinghua University, Beijing, 100084, China
} \\
[0.25em]
{\footnotesize\normalfont\ttfamily
\{zhengcj24, xcy22\}@mails.tsinghua.edu.cn
} \\
[-0.1em]
{\footnotesize\normalfont\ttfamily
\{bliang, yangjun603\}@tsinghua.edu.cn, zcs@mail.tsinghua.edu.cn
}
}

\begin{document}
\maketitle
\renewcommand{\thefootnote}{\fnsymbol{footnote}}
\footnotetext[1]{These authors contributed equally to this work.}
\footnotetext[2]{Corresponding authors.}
\renewcommand{\thefootnote}{\arabic{footnote}}

\begin{abstract}
Self-evolving LLM-based agents improve mainly by changing their agent harness: the structured execution layer around a base model, including prompts, memory, tools, middleware, runtime state, and the model--tool interaction loop.
Existing evaluations often reduce this process to isolated task scores or a single sequential curve, obscuring whether an update produces reusable improvement, overfits recent tasks, increases cost, or harms older behavior.
We introduce \textsc{SEAGym}, an evaluation environment for measuring agent harness updates across training, validation, test, replay, and cost records.
\textsc{SEAGym} turns Harbor-compatible benchmarks into dynamic self-evolution task sources with train batches, frozen update-validation, held-out ID and OOD transfer views, replay diagnostics, and saved snapshot and metric records.
Instantiating \textsc{SEAGym} on Terminal-Bench 2.0 and HLE, we compare ACE, TF-GRPO, and AHE under a shared epoch/batch protocol.
The results show that these evaluation views provide complementary signals about the evolution process: frequent updates may fail to improve held-out performance, useful intermediate snapshots may collapse later, and source diversity and model backend can affect harness reliability.
\end{abstract}

\section{Introduction}

LLM-based agents are no longer fixed systems at deployment time.
They can store experience, revise prompts, write memories, add skills, change tool-use routines, or edit runtime configuration.
These components form an \emph{agent harness}: the structured execution layer surrounding a base model, comprising prompts, context management, memory, tools, orchestration logic, middleware, runtime environments, and feedback or verification mechanisms \citep{yang2024sweagent,wang2025openhands,li2026agentharnesssurvey,pan2026nlah}.
In this paper, a \emph{self-evolving agent} is an LLM-based agent that uses task experience to update this persistent harness state, and then reuses the updated state on later tasks.

Self-evolution can occur through different processes.
An agent may update itself while solving a single task, use what it learned on one task for the next task, or repeatedly train on a set of tasks across multiple rounds.
Methods also differ in what they update.
Text-centered methods revise prompts, reflections, instructions, or experience libraries \citep{shinn2023reflexion,madaan2023selfrefine,agrawal2026gepa,youtu2025tfgrpo}.
Memory-based and skill-based methods build reusable memories, skills, workflow traces, or knowledge bases \citep{wang2023voyager,wang2024awm,xu2025amem,tang2025agentkb}.
Harness-level methods change broader execution structure, including tools, middleware, sub-agents, workflows, or project files \citep{zhuge2024gptswarm,zhang2025aflow,yuan2025evoagent,lin2026ahe}.
These differences can be expressed with the same interaction loop: the agent acts in task episodes, observes trajectories and verifier feedback, updates persistent harness state, and acts again.
Evaluating self-evolution therefore requires more than reporting whether a final agent scores higher on a task set.
The benchmark must measure the update process itself: what evidence drives each update, when snapshots improve or regress, whether improvements persist beyond the update source, and what cost or instability the update introduces.

Existing evaluations only partially support this kind of analysis.
Most agent benchmarks are designed for static evaluation: each task is an isolated episode, the agent state is reset, and the score measures one fixed agent \citep{jimenez2024swebench,zhou2024webarena,xie2024osworld,mialon2023gaia,yan2024bfcl,yao2024taubench}.
This removes the state persistence that self-evolving agents are meant to use.
Sequential and lifelong evaluations move beyond isolated episodes by studying agents over task streams \citep{jiang2026seaeval,zheng2025lifelongagentbench}.
However, other self-evolution settings remain under-investigated, such as single-task or epoch-level evolution.
More fine-grained analyses, including forgetting and regression, are also not fully covered.
A benchmark for self-evolving agents should make these settings and assessment signals explicit so that different self-evolution mechanisms can be compared under a common environment.

We introduce \textsc{SEAGym}, an evaluation environment for self-evolving LLM-based agents.
\textsc{SEAGym} uses an RL-style environment formulation in which the self-evolving agent supplies both the task policy and the harness-update rule, while the environment defines the task sampling, feedback, schedules, and snapshot assessments.
Concretely, \textsc{SEAGym} converts static benchmarks into reusable task sources, organizes them into train batches and frozen evaluation views, records agent snapshots and metric artifacts, and connects diverse methods through a rollout/update interface without prescribing how an agent updates its harness.
It represents different self-evolution processes with explicit schedule parameters, including state reset, task reuse, batch size, and update timing, so single-task adaptation, online transfer, and epoch-based batch learning can be studied under one environment.
For execution, \textsc{SEAGym} builds on Harbor, a framework for running agent evaluations and RL environments in containerized task settings \citep{Harbor_Framework}.
The two systems are complementary: Harbor provides task execution, environments, verifiers, and parallel jobs, while \textsc{SEAGym} turns static benchmark tasks into train batches, validation views, final ID and OOD transfer views, and replay diagnostics for self-evolution studies.
The experiments instantiate this path with Terminal-Bench 2.0 \citep{merrill2026terminalbench} and HLE \citep{phan2025hle}, and separate task rollout from method update so diverse self-evolving agents can be connected through thin wrappers while preserving their native update rules.

Our contributions are:
\begin{itemize}
  \item We introduce \textsc{SEAGym}, a unified evaluation environment that converts existing agent benchmarks into dynamic self-evolution task sources and supports the evaluation of self-evolving agents under a common protocol.
  \item We formulate self-evolution as an RL-style environment over agent snapshots, with configurable schedules for single-task adaptation, online transfer, and epoch-based batch learning, and with held-out views for update-validation, ID transfer, OOD transfer, replay, and diagnostics. 
  \item Through experiments on Terminal-Bench 2.0 and HLE, we show that current self-evolving mechanisms produce different update dynamics: validation gains do not always transfer, useful intermediate snapshots can regress or recover, and batch size, source diversity, and rollout backend affect harness reliability.
\end{itemize}

\section{Related Work}
\label{sec:related-work}

\paragraph{Agent harness.}
Agent harness is commonly described as structured execution layers around a base model: it comprises prompts and context management, memory, tool interfaces, orchestration logic, runtime isolation, feedback handling, tracing, and recovery logic.
Early agent work studied reasoning--acting loops and API/tool use \citep{yao2023react,schick2023toolformer,patil2023gorilla,qin2024toolllm}; newer work shows that tool documentation, agent-computer interfaces, and memory management are themselves performance-critical design variables \citep{yuan2025easytool,yang2024sweagent,xiong2025memorymanagement}.
Recent platform and protocol work further systematizes the harness as a composable agent runtime for interoperability, observability, verification, and runtime enforcement \citep{wang2025openhands,ehtesham2025interoperability,wang2026agentspec,li2026agentharnesssurvey}.
The same shift appears in production practice, where OpenAI, Anthropic, and LangChain describe harness engineering in terms of structured environments, feedback loops, durable execution, middleware, and agent-legible state \citep{openai2026harness,anthropic2026harness,langchain2025harnesses,langchain2026middleware}.
\citet{pan2026nlah} make this separation explicit by expressing harness modules in natural language, supporting inspection and ablation of the non-model agent state.
This perspective makes harness a natural target for adaptation: prompts, memories, tools, and middleware can persist beyond a single episode and affect performance on subsequent tasks.

\paragraph{Continual learning and self-evolution.}
Continual learning studies systems that face a sequence of tasks or data distributions, where learning from new experience can improve future behavior but may also interfere with previously acquired capabilities.
Its central evaluation concerns---adaptation, transfer, retention, replay, and forgetting---are therefore relevant to self-evolving agents \citep{parisi2019continual,robins1995catastrophic,kirkpatrick2017ewc,lopezpaz2017gem,chaudhry2019tiny,vandeven2019three}.
Self-evolving LLM-based agents extend this setting from parameter learning to agent-system learning.
One direction improves the underlying model or behavior policy with supervised tuning, reinforcement learning, self-play, process-level rewards, or tool-use training \citep{yuan2024selfrewarding,kumar2024score,setlur2025rewarding,choudhury2025agentprm,wang2025ragen,feng2025retool}.
A second direction treats the harness itself as the object of evolution: environmental feedback can revise prompts, memories, tool interfaces, workflows, communication structure, or middleware without retraining the base model \citep{fang2025selfevolving,zhuge2024gptswarm,zhang2025aflow,wang2024awm,yuan2025evoagent,xu2025amem,tang2025agentkb}.
This distinction matters for evaluation because harness updates are persistent, method-specific, and may be applied inside the same execution loop that produces the evidence used for updating.
Recent methods differ not only in what component they change, but also in whether they rely on reflection, verifier feedback, rollout comparison, or search, and whether the update is applied within a task, between tasks, after a batch, or over repeated epochs \citep{zhang2026ace,agrawal2026gepa,youtu2025tfgrpo,lin2026ahe}.
As a result, reported gains are hard to compare without a shared protocol that separates training episodes, validation evidence, held-out transfer, replay, and reset assumptions.

\paragraph{Agent benchmarks.}
Agent benchmarks provide the environments in which harness behavior becomes observable.
Recent surveys cover evaluations of agent capabilities and task settings such as planning, tool use, memory, software repair, terminal operation, scientific reasoning, web and desktop interaction, function calling, and tool-agent-user workflows \citep{yehudai2026agenteval,jimenez2024swebench,merrill2026terminalbench,phan2025hle,zhou2024webarena,xie2024osworld,mialon2023gaia,yan2024bfcl,yao2024taubench}.
These benchmarks mark a shift from text-only scoring toward interactive environments with executable actions, state changes, and task-specific verifiers.
Their standard protocol, however, still evaluates a fixed agent on independent episodes; persistent harness state is reset or not supported.
Benchmarks explicitly targeting self-evolution or lifelong agent learning remain scarce.
SEA-Eval evaluates agents over constructed sequential task streams and separates genuine evolution from token-consumption artifacts, while LifelongAgentBench builds skill-grounded lifelong-learning tasks across interactive database, operating-system, and knowledge-graph environments \citep{jiang2026seaeval,zheng2025lifelongagentbench}.
These efforts leave open a complementary direction: converting existing agent benchmarks into environments for both evolution and assessment, with support for specifying task sampling, feedback visibility, update timing, held-out validation, transfer tests, retention checks, and diagnostics across different self-evolution methods.

\section{Method: SEAGym}
\label{sec:framework}

\subsection{Problem Formulation}

\begin{figure*}[t]
\centering
\includegraphics[width=\textwidth]{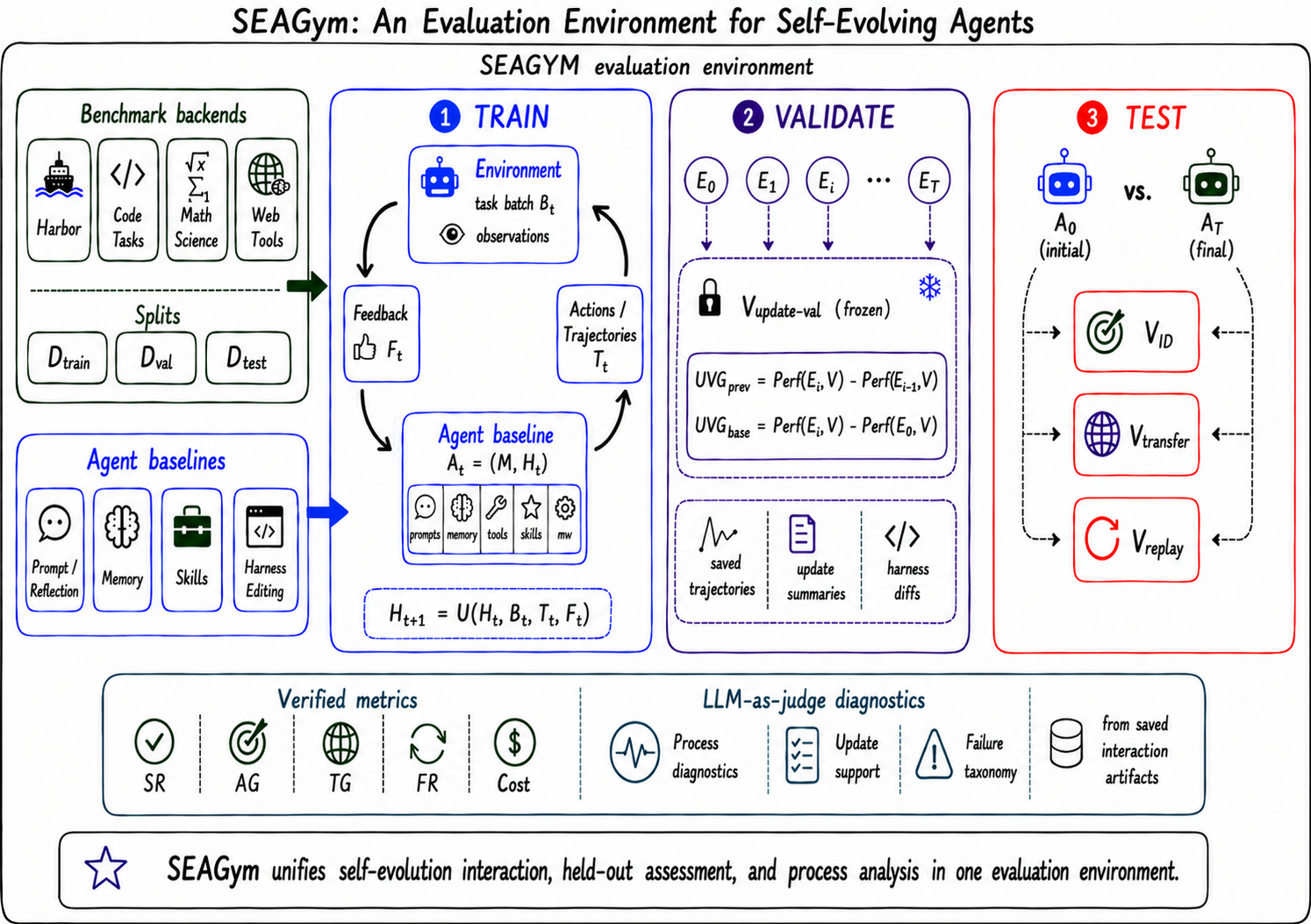}
\caption{Overview of \textsc{SEAGym}. The environment samples train batches, runs task episodes, records trajectories and verifier feedback, lets the self-evolving agent update its own state, and records evaluation points as frozen snapshots. Snapshot quality is measured with frozen update-validation, final held-out tests, replay diagnostics, and cost metrics.}
\label{fig:framework}
\end{figure*}

We model a self-evolution run as an MDP-style evaluation process
\begin{equation}
\mathcal{M}=(\mathcal{S},\mathcal{A},P,R,\rho),
\end{equation}
where each state $s_t\in\mathcal{S}$ contains the current agent snapshot, the schedule position, and the available task context.
An agent snapshot is
\begin{equation}
A_t=(M,H_t),
\end{equation}
where $M$ denotes the fixed base model and immutable runtime components, and $H_t$ denotes the mutable harness state within this execution layer: prompts, memories, skills, experience libraries, tools, middleware, project files, runtime configuration, or other model-external components used by the agent loop.
At step $t$, the evaluation environment samples a task batch $B_t$.
The agent solves the tasks, producing trajectories $\mathcal{T}_t$ and receiving feedback $F_t$ according to the benchmark's visibility policy.
It then applies its own update rule
\begin{equation}
H_{t+1}=U(H_t,B_t,\mathcal{T}_t,F_t),
\end{equation}
which, together with task execution and verifier feedback, induces the transition from $s_t$ to $s_{t+1}$.
\textsc{SEAGym} specifies the observed environment: the task distribution, feedback, schedules, and evaluation views.
It leaves the policy and update rule to each self-evolving agent and asks whether later snapshots improve beyond the update-bearing tasks.

\subsection{Evolution Schedule}

\textsc{SEAGym} represents online, single-task, batch, and epoch-based self-evolution settings through schedule parameters rather than separate benchmark formats.
This choice is important because existing self-evolving agents do not share a single natural update unit.
Some methods revise state after every task, some aggregate several trajectories before updating, and others repeatedly revisit the same source pool across epochs.
Hard-coding one schedule would therefore confound the method being evaluated with an arbitrary exposure pattern.
Instead, \textsc{SEAGym} treats state persistence, task reuse, batch size, update repeats, and assessment timing as experimental variables.
This makes it possible to ask whether a gain comes from the update rule itself, from seeing more diverse evidence in one update, from performing more frequent updates, or from repeatedly revisiting the same task distribution.
The main experiments use persistent state, repeated train batches, train-batch updates, and epoch-end frozen update-validation assessment.
Appendix~\ref{app:protocol-details} gives the schedule fields, saved records, and epoch loop in Algorithm~\ref{alg:seagym}.

\subsection{Data Splits and Evaluation Views}

\textsc{SEAGym} separates dataset splits from evaluation views.
The base manifest keeps the conventional split structure $D_{\text{train}}$, $D_{\text{val}}$, and $D_{\text{test}}$, which controls task visibility and update evidence.
Evaluation views are materialized from these splits to assess different aspects of the self-evolution process: update-validation tracks intermediate snapshots, ID and OOD views test held-out transfer at different distribution distances, and replay views measure retention, forgetting, or regression.
Table~\ref{tab:splits} summarizes the resulting pools and views.

\begin{table*}[t]
\centering
\small
\renewcommand{\arraystretch}{1.08}
\begin{tabular}{@{}p{0.16\linewidth}p{0.20\linewidth}p{0.58\linewidth}@{}}
\toprule
\textbf{Role} & \textbf{Pool/view} & \textbf{Question answered}\\
\midrule
Update evidence & $B_t\subset D_{\text{train}}$ & what trajectories and visible feedback drive the next harness update?\\
Process assessment & $V_{\text{update-val}}\subset D_{\text{val}}$ & does a frozen intermediate snapshot improve without using validation tasks as update evidence?\\
ID transfer & $V_{\text{ID}}\subset D_{\text{test}}$ & does the update transfer to unseen tasks from the source distribution?\\
OOD transfer & $V_{\text{OOD}}\subset D_{\text{test}}$ & does the update transfer to shifted or target-domain tasks?\\
Replay / retention & $V_{\text{replay}}$ & does the updated harness preserve or recover earlier behavior?\\
\bottomrule
\end{tabular}
\caption{Dataset splits and evaluation views in \textsc{SEAGym}. Base splits control task visibility and update evidence, while views define assessment lenses over the self-evolution process.}
\label{tab:splits}
\end{table*}

The main protocol evaluates native self-updates without using validation views to modify them.
Train batches produce update evidence, while validation and test views are reserved for frozen assessment.
If validation examples or private verifier artifacts were mixed into the update evidence, a higher score could reflect direct adaptation to the assessment view rather than self-evolution that transfers to future tasks.
Conversely, if only the final test were reported, the evaluation would hide whether an intermediate snapshot improved, regressed, recovered, or merely became more expensive.
\textsc{SEAGym} therefore records saved snapshot and metric records at explicit evaluation points: update-validation tracks the process, final ID and OOD views test held-out transfer at different distribution distances, and replay views expose retention or regression.

\subsection{Benchmark and Baseline Integration}

\textsc{SEAGym} is designed to integrate both benchmark suites and self-evolving methods with minimal code adaptation.
On the benchmark side, it reuses Harbor as the task runner \citep{Harbor_Framework}.
Harbor retains task definitions, environments, verifiers, and parallel execution, while \textsc{SEAGym} adds the self-evolution schedule: train batches, frozen validation views, final held-out views, snapshot timing, and metric computation.
This turns static benchmark tasks into a dynamic self-evolution environment without rewriting the benchmark.

On the method side, \textsc{SEAGym} separates rollout from update.
A rollout component runs tasks under the current harness state and returns a trajectory batch.
An update component consumes that trajectory batch and applies the method's native self-evolution rule.
This decomposition gives prompt, memory, skill, textual-optimization, context-update, and harness-editing methods a common interface while leaving each method's update semantics intact.
In practice, connecting a new baseline only requires a thin wrapper that converts \textsc{SEAGym} trajectory batches to the method's native update input and saves the resulting harness state.
The same boundary also applies to reusable integration checklists and skill templates: benchmark authors follow Harbor task/adapter templates, and baseline authors implement the rollout/update wrapper.
Additional interface details are in Appendix~\ref{app:integration-details}.
Figure~\ref{fig:framework} summarizes this flow.

\subsection{Metrics}

\textsc{SEAGym} computes metrics from saved records so results can be inspected and recomputed without rerunning environments.
Given verifier score $r(A,x)\in[0,1]$, it reports
\begin{align}
\text{Perf}(A,D)&=\frac{1}{|D|}\sum_{x\in D} r(A,x),\\
\text{SR}(A,D)&=\frac{1}{|D|}\sum_{x\in D}\mathbb{I}[r(A,x)=1].
\end{align}
For update-validation view $V=V_{\text{update-val}}$ and evaluation point $E_i$:
\begin{align}
\text{UVG}^{\text{prev}}_i &= \text{Perf}(E_i,V) - \text{Perf}(E_{i-1},V),\\
\text{UVG}^{\text{base}}_i &= \text{Perf}(E_i,V) - \text{Perf}(E_0,V).
\end{align}
For compact notation, let $D_I$, $D_O$, and $D_R$ denote the task sets underlying the final ID, OOD, and replay views:
\begin{align}
\text{IDG} &= \text{Perf}(A_T,D_I) - \text{Perf}(A_0,D_I),\\
\text{OODG} &= \text{Perf}(A_T,D_O) - \text{Perf}(A_0,D_O),\\
\text{FR} &= \max(0,\text{Perf}(A_0,D_R) - \text{Perf}(A_T,D_R)).
\end{align}
We also report token usage, tool calls, wall-clock time, and cost reduction when available.
Main result tables use domain-level macro averages by default so that larger task groups do not dominate cross-domain conclusions.
Appendix~\ref{app:metric-details} summarizes the metric table and additional details.

\subsection{Saved Records and Process Diagnostics}

Primary \textsc{SEAGym} scores are computed from verified task rewards, held-out evaluation views, replay checks, and measured cost.
However, these metrics alone do not fully explain why a self-evolution run improves, regresses, recovers, or becomes more expensive.
\textsc{SEAGym} therefore saves trajectory references, public feedback, update summaries, harness diffs when available, snapshot records, and metric records at each evaluation point.
These artifacts support offline process diagnostics, such as whether an update is supported by the observed trajectories, whether failures come from task strategy or runtime behavior, and whether a later snapshot recovers from an earlier regression.

Such diagnostics are secondary to the verified metrics.
They may be produced by manual inspection, rule-based analysis, or optional offline LLM-as-judge annotators.
This separation keeps the primary evaluation tied to executable task outcomes while still making the evolution process interpretable.

\section{Experiments}
\label{sec:experiments}

\subsection{Experimental Setup}
\label{sec:experiments-setup}

We use an epoch-based batch setting over 80 source train tasks from Terminal-Bench 2.0 and HLE text-only Math/Physics, with 35 source validation tasks, 55 source test tasks, and 80 HLE CS/AI and Engineering OOD transfer tasks \citep{merrill2026terminalbench,phan2025hle}.
Unless stated otherwise, runs use five epochs and train batch size 20.
The main comparison evaluates ACE, TF-GRPO, and AHE under DeepSeek-V4-Flash; the ablations use AHE for batch size, source diversity, and cross-model transfer \citep{zhang2026ace,youtu2025tfgrpo,lin2026ahe,deepseekai2026deepseekv4api,openai2026gpt54,zai2026glm51}.
Appendix Tables~\ref{tab:paper-experiment-settings} and~\ref{tab:experiment-specific-configs} list the full settings, and Appendix~\ref{app:metric-details} summarizes the metric records.
In result tables, $V_i$, ID$_i$, and OOD$_i$ denote success rates of snapshot $i$ on the update-validation, in-distribution transfer, and OOD transfer views, respectively; subscript $0$ denotes the initial snapshot, $T$ the final snapshot, and $\star$ the best-validation snapshot selected for reporting.

\subsection{Baseline Results}
\label{sec:experiments-baseline-results}

\begin{figure}[tbp]
\centering
\includegraphics[width=0.92\linewidth]{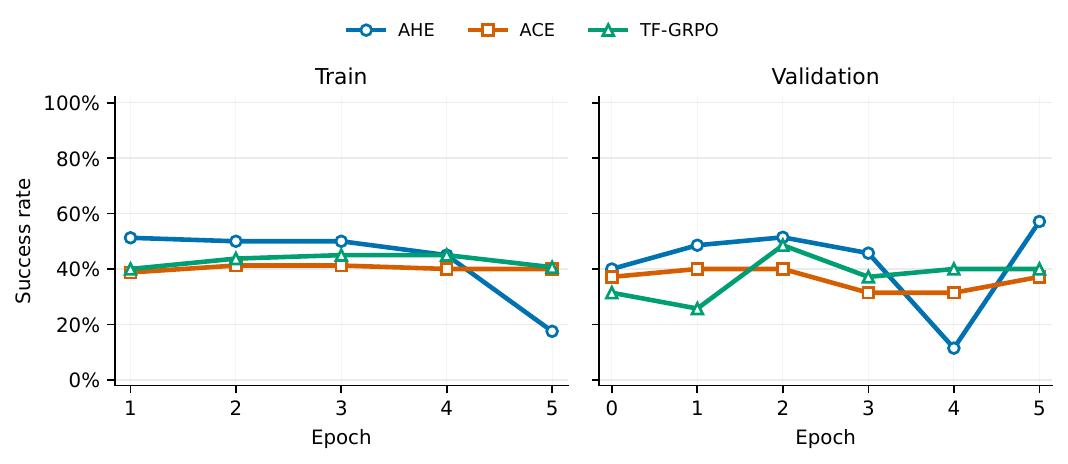}
\caption{Baseline learning curves for AHE, ACE, and TF-GRPO.}
\label{fig:baseline-results-curves}
\end{figure}

Figure~\ref{fig:baseline-results-curves} and Table~\ref{tab:baseline-results} compare the three DeepSeek-V4-Flash runs.
Appendix Figures~\ref{fig:baseline-results-breakdown-curves} and~\ref{fig:baseline-results-batch-index-curves} provide source-group and batch-index breakdowns.

\begin{table*}[tbp]
\centering
\scriptsize
\setlength{\tabcolsep}{3.2pt}
\resizebox{\textwidth}{!}{%
\begin{tabular}{@{}lrrrrrrrrrrr@{}}
\toprule
\textbf{Agent} & \textbf{$V_0$} & \textbf{$V_\star$} &
\textbf{UVG} & \textbf{ID$_0$} & \textbf{ID$_\star$} &
\textbf{IDG} & \textbf{OOD$_0$} & \textbf{OOD$_\star$} & \textbf{OODG} &
\textbf{Rollout tok./task} & \textbf{Update tok./update}\\
\midrule
AHE & 40.0 & 57.1 & +17.1 & 40.0 & 49.1 & +9.1 & 22.5 & 28.8 & +6.3 & 1.46M & 3.91M\\
ACE & 37.1 & 40.0 & +2.9 & 30.9 & 34.5 & +3.6 & 22.5 & 25.0 & +2.5 & 1.93M & --\\
TF-GRPO & 31.4 & 48.6 & +17.1 & 30.9 & 34.5 & +3.6 & 26.3 & 23.8 & -2.5 & 2.33M & 1.60M\\
\bottomrule
\end{tabular}
}
\caption{Baseline results. Success rates are percentages, gains are percentage points, and token costs are normalized per task or update. All methods report the best-validation snapshot; for AHE, this selected snapshot is the final snapshot.}
\label{tab:baseline-results}
\end{table*}

AHE is the only baseline that improves validation, ID, and OOD together.
This pattern is easier to interpret when we distinguish what each method optimizes.
AHE changes the agent harness itself, including prompts, tool-use constraints, middleware, and runtime behavior.
Its updates can therefore alter how the agent searches for evidence, validates candidate answers, recovers from tool errors, and decides when to stop.
This broader editable scope helps explain why AHE transfers best across the three reported views, but it also creates a larger reliability burden: a harmful harness change can affect many otherwise unrelated tasks.

ACE instead behaves more like skill- or strategy-memory optimization.
It yields modest ID and OOD gains, suggesting that the learned skillbook provides reusable task-handling knowledge.
However, because ACE does not directly rewrite the execution path or the tool/middleware contract, it has less leverage over failures caused by interaction policy, environment handling, or runtime behavior, and its validation gain is correspondingly smaller.
TF-GRPO lies between these two patterns: grouped rollout evidence can quickly strengthen behavior on the source distribution, which produces a large validation gain and a small ID gain, but the OOD drop and highest rollout cost suggest that this adaptation does not reliably transfer to shifted target tasks.
Thus, validation gain and update activity alone are insufficient; the benchmark separates what is being optimized, whether the change generalizes to held-out views, and what cost or instability the update mechanism introduces.

\subsection{Training Fix and Forgetting}
\label{sec:experiments-fix-forgetting}

\begin{figure*}[t]
\centering
\includegraphics[width=1\textwidth]{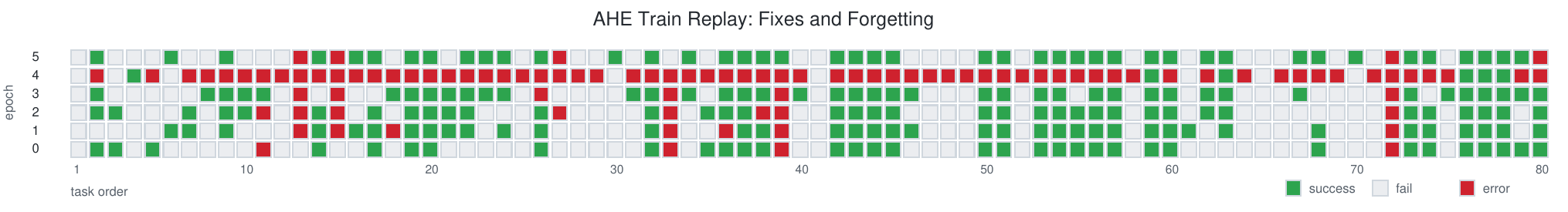}
\caption{AHE train replay grid. Green cells are successful trials, gray cells are failed trials, and red cells are rollout errors.}
\label{fig:fix-forgetting-grid}
\end{figure*}

\begin{figure*}[t]
\centering
\includegraphics[width=0.98\textwidth]{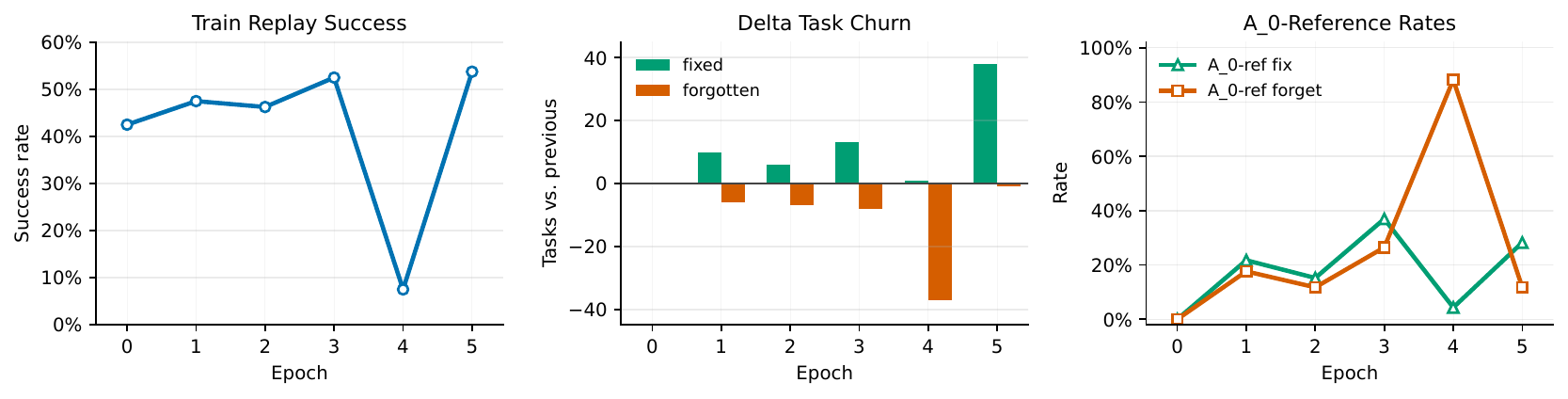}
\caption{AHE train replay diagnostics. Left: success rate on the 80 source train tasks. Middle: pairwise delta task churn, where fixed tasks are newly solved relative to the previous snapshot and forgotten tasks are previously solved tasks that fail at the current snapshot. Right: $A_0$-reference fix/forget rates with fixed denominators.}
\label{fig:fix-forgetting-summary}
\end{figure*}

We next replay the source train set with AHE at initialization, after each epoch, and at the end of training.
Figure~\ref{fig:fix-forgetting-grid} reports the task-level replay grid.
Figure~\ref{fig:fix-forgetting-summary} reports the replay success rate, pairwise delta task churn, and $A_0$-reference fix/forget rates; Appendix Figure~\ref{fig:fix-forgetting-by-source} provides the source-group breakdown.

The final agent solves 43/80 train-replay tasks, compared with 34/80 for the initial agent, but the trajectory is not monotonic.
After epoch 4, replay performance drops to 6/80 and produces many rollout errors before the final agent recovers.
This shows why snapshot-level diagnostics are needed: an initial-versus-final table would miss a damaging intermediate harness regression and would not explain where the variance comes from.
Relative to $A_0$, the final agent fixes 13 initially failed tasks and forgets 4 initially solved tasks, giving a net gain of 9 train-replay tasks.
The process-level view is the main point of this diagnostic.
Early epochs add useful harness behavior, such as more active evidence gathering, stricter answer checks, tool-error recovery, and completion guards; these changes raise the fix count, but they also introduce new middleware constraints and execution paths that break some previously solved tasks.
After epoch 4, the dominant failure is an evolved message-construction regression in the middleware/runtime contract; the final epoch restores that execution path, so many tasks recover at once and the $A_0$-reference forget rate falls.
Thus, replay does not merely report a final retention score: it decomposes self-evolution variance into task-behavior churn and execution-path instability, which is precisely the kind of process diagnostic \textsc{SEAGym} is designed to expose.
Appendix~\ref{app:fix-forgetting-metrics} gives the metric definitions, and Appendix~\ref{app:training-forgetting-case-study} analyzes the observed update sequence.

\subsection{Effect of Batch Size}
\label{sec:experiments-batch-size}

We vary the AHE train batch size while keeping the train, update-validation, and ID test task sets fixed.
Table~\ref{tab:batch-size} summarizes validation, ID, cost, and update status; Appendix Figures~\ref{fig:batch-size-breakdown-curves} and~\ref{fig:batch-size-batch-index-curves} provide curve-level breakdowns, and Appendix~\ref{app:batch-size-case-study} analyzes the corresponding update artifacts.

\begin{center}
\centering
\scriptsize
\setlength{\tabcolsep}{2.0pt}
\begin{tabular}{@{}rccrr@{}}
\toprule
\textbf{Batch} & \textbf{Validation} & \textbf{ID test} &
\textbf{Update tok.} & \textbf{Updates}\\
\midrule
10 & 37.1$\to$22.9 (-14.3) & 38.2$\to$23.6 (-14.5) & 3.13M & 39/40\\
20 & 40.0$\to$57.1 (\textbf{+17.1}) & 40.0$\to$49.1 (\textbf{+9.1}) & 3.91M & 20/20\\
40 & 37.1$\to$40.0 (\textbf{+2.9}) & 41.8$\to$43.6 (\textbf{+1.8}) & 3.36M & 10/10\\
80 & 42.9$\to$25.7 (-17.1) & 41.8$\to$25.5 (-16.4) & 3.57M & 5/5\\
\bottomrule
\end{tabular}
\captionof{table}{Effect of batch size. Validation and ID columns show initial $\rightarrow$ final success rate, with gain in parentheses. Update tokens are normalized per SEAGym update call; rollout token costs are reported in Appendix~\ref{app:experimental-details}.}
\label{tab:batch-size}
\end{center}

Batch size is non-monotonic: batch 20 is the only setting with large positive validation and ID gains, while batch 10 and batch 80 regress.
This suggests that AHE is limited by how much evidence the evolving agent can analyze in a single update, not only by update frequency.
The recorded update cost is similar across batch sizes, roughly 3--4M tokens per update, so increasing the batch size does not give the evolving agent proportionally more analysis capacity.
At batch 80, the same update must inspect too many trajectories, diluting per-task attention and increasing the risk of broad, brittle middleware changes.
At batch 10, each update sees too little evidence and the run performs twice as many harness updates as batch 20, making the update stream higher-variance and giving runtime regressions more opportunities to accumulate.
The middleware/runtime failures observed in the batch-10 and batch-80 runs are therefore not merely external noise; they are part of the schedule-induced instability exposed by the benchmark.
Neither larger batches nor more frequent updates are automatically better; in this sweep, batch 20 is the setting where evidence diversity, per-task analysis depth, update frequency, and harness stability are most balanced.

\subsection{Source Diversity}
\label{sec:experiments-source-diversity}

We compare the main mixed-source AHE run with an HLE-only run of the same train size.
Tables~\ref{tab:source-diversity-final} and~\ref{tab:source-diversity-selected} report final reliability and the best available intermediate snapshot; Appendix Figures~\ref{fig:source-diversity-breakdown-curves} and~\ref{fig:source-diversity-batch-index-curves} provide curve-level breakdowns, and Appendix~\ref{app:source-diversity-case-study} analyzes the corresponding update artifacts.

\begin{center}
\centering
\scriptsize
\setlength{\tabcolsep}{6.0pt}
\begin{tabular}{@{}lrrrrr@{}}
\toprule
\textbf{Source} & \textbf{$V_T$} & \textbf{ID$_T$} & \textbf{IDG$_T$} & \textbf{OOD$_T$} & \textbf{OODG$_T$}\\
\midrule
Terminal-Bench + HLE & 57.1 & 49.1 & +9.1 & 28.8 & +6.3\\
HLE only & 0.0 & 0.0 & -22.0 & 0.0 & -21.2\\
\bottomrule
\end{tabular}
\captionof{table}{Source diversity at the final snapshot. Success rates are percentages and gains are percentage points.}
\label{tab:source-diversity-final}
\end{center}

\begin{center}
\centering
\scriptsize
\setlength{\tabcolsep}{6.0pt}
\begin{tabular}{@{}lrrrrr@{}}
\toprule
\textbf{Source} & \textbf{Epoch$_\star$} & \textbf{ID$_\star$} & \textbf{IDG$_\star$} & \textbf{OOD$_\star$} & \textbf{OODG$_\star$}\\
\midrule
Terminal-Bench + HLE & 5 (final) & 49.1 & +9.1 & 28.8 & +6.3\\
HLE only & 3 & 47.3 & +7.3 & 25.0 & +3.8\\
\bottomrule
\end{tabular}
\captionof{table}{Source diversity at the selected epoch. The selected epoch is the best available intermediate snapshot. Success rates are percentages and gains are percentage points.}
\label{tab:source-diversity-selected}
\end{center}

The HLE-only run reaches a useful intermediate snapshot, but its final snapshot collapses on validation, ID, and OOD.
This suggests that a single benchmark can drive the harness toward benchmark-specific local optima.
The mixed-source run also passes through a bad intermediate state, so diversity does not prevent harmful updates.
Its advantage is recovery: Terminal-Bench exposes tool, environment, and execution failures, while HLE exposes reasoning failures, giving later updates more varied evidence for restoring a broken harness.

\subsection{Cross-Model Transfer}
\label{sec:experiments-cross-model-transfer}

\begin{figure}[tbp]
\centering
\includegraphics[width=0.96\linewidth]{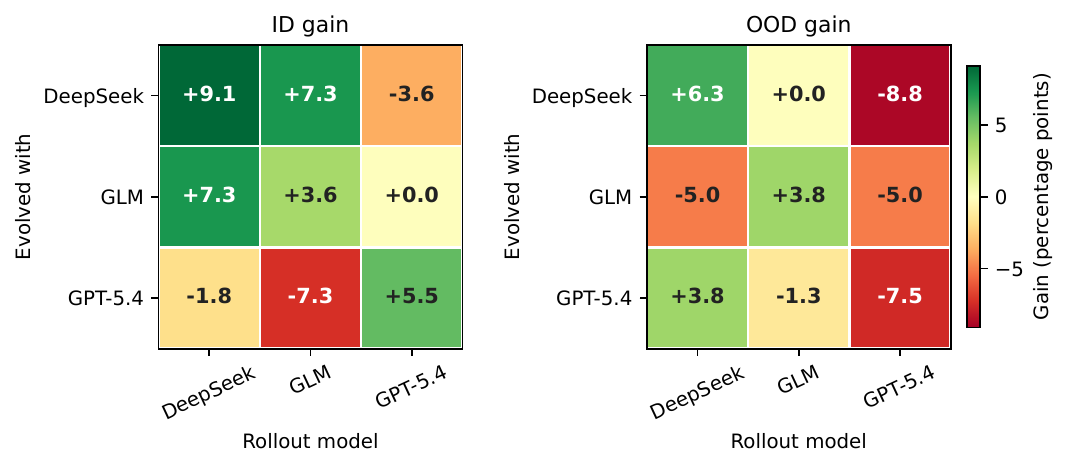}
\caption{Cross-model ID and OOD gains. Rows indicate the rollout model used to evolve the AHE harness, columns indicate the rollout model used at evaluation time, and each cell reports gain over the same rollout model's initial harness on the same evaluation set. Values are percentage points.}
\label{fig:cross-model-transfer-heatmap}
\end{figure}

Finally, we swap evolved AHE harnesses across rollout models.
Figure~\ref{fig:cross-model-transfer-heatmap} reports ID and OOD gains for the best-validation AHE snapshot from each rollout backend.
Appendix~\ref{app:cross-model-case-study} analyzes the update artifacts behind these results, and Appendix~\ref{app:cross-model-continuation} reports the corresponding training trajectories and full success-rate table.

The same-backend results show that AHE can adapt the harness to the rollout backend that produced the updates: ID gains are positive for all three backends, ranging from +3.6 to +9.1 percentage points.
Cross-backend results are less stable and asymmetric.
For example, the DeepSeek-evolved harness improves GLM ID by +7.3 points but hurts GPT-5.4 ID by -3.6 points, while the GPT-5.4-evolved harness improves GPT-5.4 ID by +5.5 points but hurts GLM ID by -7.3 points.
The update artifacts explain this asymmetry.
DeepSeek trajectories lead AHE to edit verification, tool-recovery, artifact-cleanup, and message-contract paths; GLM trajectories emphasize text-only reasoning and research without output; GPT-5.4 trajectories emphasize artifact constraints and validation sufficiency.
Harness gains therefore transfer when the evaluation trajectory exposes similar failures, and weaken when the edited subsystem no longer matches the dominant failure surface.

ID gains also do not imply OOD gains.
GPT-5.4-evolved AHE improves GPT-5.4 ID by +5.5 points but drops GPT-5.4 OOD by -7.5 points, and most cross-backend OOD gains are neutral or negative.
This mismatch shows that an update can fit the training backend's observed interaction failures while failing to cover the shifted failure modes of another rollout backend or target domain.
Thus, if we only reported native ID results, all three backends would appear to improve; the cross-model and OOD views reveal when the learned harness changes remain aligned with the evaluation trajectories and when they do not.
\textsc{SEAGym}'s separate ID, OOD, and cross-backend assessments make this difference visible instead of collapsing it into a single final score.

\section{Conclusion}

\textsc{SEAGym} evaluates self-evolving LLM agents by treating harness change as the object of study.
The central question is not only whether the final agent scores higher, but what persistent state is updated, when the update helps, whether the improvement transfers beyond the update source, whether earlier behavior is lost, and what execution or update cost is required.
To support this view, \textsc{SEAGym} converts Harbor-compatible benchmarks into train batches, frozen update-validation views, held-out ID and OOD transfer views, replay diagnostics, and saved snapshot and metric records, while allowing each method to keep its native update rule.

The Terminal-Bench 2.0 and HLE experiments show that self-evolution gains are strongly tied to the update mechanism.
AHE edits the harness itself and can produce broad validation, ID, and OOD gains, but its larger editable scope also creates process-level reliability risks.
ACE behaves more like skill- or strategy-memory optimization: it accumulates reusable task-handling knowledge, but has less leverage over failures caused by runtime behavior or environment interaction.
TF-GRPO strengthens behavior from grouped rollout evidence, yielding strong source-validation gains, but its OOD drop and rollout cost show that such adaptation need not transfer stably.

These findings argue against reducing self-evolution to a single final score or validation curve.
Useful intermediate states can later regress, source diversity and batch size can change harness reliability, and model backend can condition whether an evolved harness transfers.
\textsc{SEAGym} is therefore intended not as a one-shot ranking of self-evolving methods, but as a controlled evaluation environment with saved records for comparing what different methods update, whether those updates generalize, and what costs or instabilities they introduce.

\section*{Limitations}

This instantiation of \textsc{SEAGym} focuses on harness- and state-level self-evolution on Terminal-Bench 2.0 and HLE.
These sources cover complementary execution-heavy and reasoning-heavy settings, and demonstrate how the protocol separates update-validation, ID transfer, OOD transfer, replay, and cost signals.
Future work can extend the same protocol to additional agentic domains, such as web or desktop interaction, long-horizon software engineering, data-analysis workflows, multi-agent collaboration, and continuous online task streams.
Because \textsc{SEAGym} separates benchmark execution from the self-evolution schedule, such extensions mainly require new Harbor-compatible task sources and evaluation views rather than changes to the core protocol.

The experiments study model-external harness evolution: agents update prompts, memories, skills, experience context, middleware, or tool-use policies.
The same environment can be extended to model-weight updates, online RL fine-tuning, or hybrid systems, making it possible to compare harness-level and parameter-level learning in terms of cost, stability, and transfer.

The multi-view design creates a cost/coverage tradeoff because snapshots are saved and re-evaluated across update-validation, ID, OOD, and replay views.
Future work can study more efficient snapshot selection, adaptive replay, budget-aware evaluation, and more systematic process diagnostics while preserving visibility into regression, recovery, transfer, and forgetting.

Finally, the OOD and cross-model experiments show that harness updates can depend on both task distribution and rollout backend.
Expanding to more source/target domain pairs, model backends, and longer horizons would further clarify which update mechanisms produce stable transfer and which become backend- or benchmark-specific.

\section*{Ethics Statement}

\textsc{SEAGym} is intended as an evaluation framework.
The benchmark should avoid exposing hidden verifier details or private oracle artifacts to agents.
If future task domains include real user data, web data, or proprietary repositories, dataset construction must follow licensing, privacy, and anonymization requirements.

\FloatBarrier
\bibliography{custom}

@misc{Harbor_Framework,
  author = {{Harbor Framework Team}},
  month = jan,
  title = {{Harbor: A framework for evaluating and optimizing agents and models in container environments}},
  url = {https://github.com/harbor-framework/harbor},
  year = {2026}
}

@inproceedings{yao2023react,
  title = {{ReAct}: Synergizing Reasoning and Acting in Language Models},
  author = {Yao, Shunyu and Zhao, Jeffrey and Yu, Dian and Du, Nan and Shafran, Izhak and Narasimhan, Karthik and Cao, Yuan},
  booktitle = {International Conference on Learning Representations},
  year = {2023},
  url = {https://openreview.net/forum?id=WE_vluYUL-X}
}

@inproceedings{shinn2023reflexion,
  title = {Reflexion: Language Agents with Verbal Reinforcement Learning},
  author = {Shinn, Noah and Cassano, Federico and Berman, Edward and Gopinath, Ashwin and Narasimhan, Karthik and Yao, Shunyu},
  booktitle = {Advances in Neural Information Processing Systems},
  year = {2023},
  url = {https://arxiv.org/abs/2303.11366}
}

@inproceedings{madaan2023selfrefine,
  title = {Self-Refine: Iterative Refinement with Self-Feedback},
  author = {Madaan, Aman and Tandon, Niket and Gupta, Prakhar and Hallinan, Skyler and Gao, Luyu and Wiegreffe, Sarah and Alon, Uri and Dziri, Nouha and Prabhumoye, Shrimai and Yang, Yiming and Gupta, Shashank and Majumder, Bodhisattwa Prasad and Hermann, Katherine and Welleck, Sean and Yazdanbakhsh, Amir and Clark, Peter},
  booktitle = {Advances in Neural Information Processing Systems},
  year = {2023},
  url = {https://arxiv.org/abs/2303.17651}
}

@inproceedings{jimenez2024swebench,
  title = {{SWE}-bench: Can Language Models Resolve Real-World GitHub Issues?},
  author = {Jimenez, Carlos E. and Yang, John and Wettig, Alexander and Yao, Shunyu and Pei, Kexin and Press, Ofir and Narasimhan, Karthik},
  booktitle = {International Conference on Learning Representations},
  year = {2024},
  url = {https://openreview.net/forum?id=VTF8yNQM66}
}

@inproceedings{zhou2024webarena,
  title = {{WebArena}: A Realistic Web Environment for Building Autonomous Agents},
  author = {Zhou, Shuyan and Xu, Frank F. and Zhu, Hao and Zhou, Xuhui and Lo, Robert and Sridhar, Abishek and Cheng, Xianyi and Ou, Tianyue and Bisk, Yonatan and Fried, Daniel and Alon, Uri and Neubig, Graham},
  booktitle = {International Conference on Learning Representations},
  year = {2024},
  url = {https://openreview.net/forum?id=oKn9c6ytLx}
}

@article{xie2024osworld,
  title = {{OSWorld}: Benchmarking Multimodal Agents for Open-Ended Tasks in Real Computer Environments},
  author = {Xie, Tianbao and Zhang, Danyang and Chen, Jixuan and Li, Xiaochuan and Zhao, Siheng and Cao, Ruisheng and Hua, Tuo and Cheng, Zhoujun and Shin, Dongchan and Lei, Fangyu and Liu, Yiheng and Xu, Yiheng and Zhou, Shuyan and Savarese, Silvio and Xiong, Caiming and Zhong, Victor and Yu, Tao},
  journal = {arXiv preprint arXiv:2404.07972},
  year = {2024},
  url = {https://arxiv.org/abs/2404.07972}
}

@inproceedings{mialon2023gaia,
  title = {{GAIA}: A Benchmark for General {AI} Assistants},
  author = {Mialon, Gr{\'e}goire and Fourrier, Cl{\'e}mentine and Wolf, Thomas and LeCun, Yann and Scialom, Thomas},
  booktitle = {International Conference on Learning Representations},
  year = {2024},
  url = {https://openreview.net/forum?id=fibxvahvs3}
}

@inproceedings{yan2024bfcl,
  title = {The {Berkeley Function Calling Leaderboard} ({BFCL}): From Tool Use to Agentic Evaluation of Large Language Models},
  author = {Patil, Shishir G. and Mao, Huanzhi and Yan, Fanjia and Ji, Charlie Cheng-Jie and Suresh, Vishnu and Stoica, Ion and Gonzalez, Joseph E.},
  booktitle = {Proceedings of the 42nd International Conference on Machine Learning},
  pages = {48371--48392},
  volume = {267},
  series = {Proceedings of Machine Learning Research},
  publisher = {PMLR},
  year = {2025},
  url = {https://proceedings.mlr.press/v267/patil25a.html}
}

@article{yao2024taubench,
  title = {tau-bench: A Benchmark for Tool-Agent-User Interaction in Real-World Domains},
  author = {Yao, Shunyu and Shinn, Noah and Razavi, Parth and Narasimhan, Karthik},
  journal = {arXiv preprint arXiv:2406.12045},
  year = {2024},
  url = {https://arxiv.org/abs/2406.12045}
}

@article{merrill2026terminalbench,
  title = {{Terminal-Bench}: Benchmarking Agents on Hard, Realistic Tasks in Command Line Interfaces},
  author = {Merrill, William and others},
  journal = {arXiv preprint arXiv:2601.11868},
  year = {2026},
  url = {https://arxiv.org/abs/2601.11868}
}

@article{phan2025hle,
  title = {Humanity's Last Exam},
  author = {Phan, Long and others},
  journal = {arXiv preprint arXiv:2501.14249},
  year = {2025},
  url = {https://arxiv.org/abs/2501.14249}
}

@misc{deepseekai2026deepseekv4api,
  title = {{DeepSeek V4 Preview Release}},
  author = {{DeepSeek-AI}},
  year = {2026},
  howpublished = {\url{https://api-docs.deepseek.com/news/news260424}}
}

@misc{openai2026gpt54,
  title = {{OpenAI API} Model Documentation: {GPT-5.4}},
  author = {{OpenAI}},
  year = {2026},
  howpublished = {\url{https://platform.openai.com/docs/models/gpt-5.4}}
}

@misc{zai2026glm51,
  title = {{GLM-5.1} Model Documentation},
  author = {{Z.AI}},
  year = {2026},
  howpublished = {\url{https://docs.z.ai/guides/llm/glm-5.1}}
}

@article{parisi2019continual,
  title = {Continual Lifelong Learning with Neural Networks: A Review},
  author = {Parisi, German I. and Kemker, Ronald and Part, Jose L. and Kanan, Christopher and Wermter, Stefan},
  journal = {Neural Networks},
  volume = {113},
  pages = {54--71},
  year = {2019},
  doi = {10.1016/j.neunet.2019.01.012}
}

@article{jiang2026seaeval,
  title = {{SEA}-Eval: A Benchmark for Evaluating Self-Evolving Agents Beyond Episodic Assessment},
  author = {Jiang, Sihang and Ma, Lipeng and Hong, Zhonghua and Wang, Keyi and Lu, Zhiyu and Chen, Shisong and Zhang, Jinghao and Pan, Tianjun and Zhou, Weijia and Liang, Jiaqing and Xiao, Yanghua},
  journal = {arXiv preprint arXiv:2604.08988},
  year = {2026},
  url = {https://arxiv.org/abs/2604.08988}
}

@article{zheng2025lifelongagentbench,
  title = {{LifelongAgentBench}: Evaluating {LLM} Agents as Lifelong Learners},
  author = {Zheng, Junhao and Cai, Xidi and Li, Qiuke and Zhang, Duzhen and Li, ZhongZhi and Zhang, Yingying and Song, Le and Ma, Qianli},
  journal = {arXiv preprint arXiv:2505.11942},
  year = {2025},
  url = {https://arxiv.org/abs/2505.11942}
}

@inproceedings{schick2023toolformer,
  title = {Toolformer: Language Models Can Teach Themselves to Use Tools},
  author = {Schick, Timo and Dwivedi-Yu, Jane and Dess{\`i}, Roberto and Raileanu, Roberta and Lomeli, Maria and Zettlemoyer, Luke and Cancedda, Nicola and Scialom, Thomas},
  booktitle = {Advances in Neural Information Processing Systems},
  year = {2023},
  url = {https://arxiv.org/abs/2302.04761}
}

@article{patil2023gorilla,
  title = {Gorilla: Large Language Model Connected with Massive {API}s},
  author = {Patil, Shishir G. and Zhang, Tianjun and Wang, Xin and Gonzalez, Joseph E.},
  journal = {arXiv preprint arXiv:2305.15334},
  year = {2023},
  url = {https://arxiv.org/abs/2305.15334}
}

@inproceedings{qin2024toolllm,
  title = {{ToolLLM}: Facilitating Large Language Models to Master 16000+ Real-world {API}s},
  author = {Qin, Yujia and Liang, Shihao and Ye, Yining and Zhu, Kunlun and Yan, Lan and Lu, Yaxi and Lin, Yankai and Cong, Xin and Tang, Xiangru and Qian, Bill and Zhao, Sihan and Hong, Lauren and Tian, Runchu and Xie, Ruobing and Zhou, Jie and Gerstein, Mark and Li, Dahai and Liu, Zhiyuan and Sun, Maosong},
  booktitle = {International Conference on Learning Representations},
  year = {2024},
  url = {https://openreview.net/forum?id=dHng2O0Jjr}
}

@inproceedings{yuan2025easytool,
  title = {{EASYTOOL}: Enhancing {LLM}-based Agents with Concise Tool Instruction},
  author = {Yuan, Siyu and Song, Kaitao and Chen, Jiangjie and Tan, Xu and Shen, Yongliang and Kan, Ren and Li, Dongsheng and Yang, Deqing},
  booktitle = {Proceedings of the 2025 Conference of the Nations of the Americas Chapter of the Association for Computational Linguistics: Human Language Technologies},
  pages = {951--972},
  year = {2025},
  doi = {10.18653/v1/2025.naacl-long.44},
  url = {https://aclanthology.org/2025.naacl-long.44/}
}

@article{wang2023voyager,
  title = {Voyager: An Open-Ended Embodied Agent with Large Language Models},
  author = {Wang, Guanzhi and Xie, Yuqi and Jiang, Yunfan and Mandlekar, Ajay and Xiao, Chaowei and Zhu, Yuke and Fan, Linxi and Anandkumar, Anima},
  journal = {arXiv preprint arXiv:2305.16291},
  year = {2023},
  url = {https://arxiv.org/abs/2305.16291}
}

@article{wang2025openhands,
  title = {The {OpenHands} Software Agent {SDK}: A Composable and Extensible Foundation for Production Agents},
  author = {Wang, Xingyao and Rosenberg, Simon and Michelini, Juan and Smith, Calvin and Tran, Hoang and Nyst, Engel and Malhotra, Rohit and Zhou, Xuhui and Chen, Valerie and Brennan, Robert and others},
  journal = {arXiv preprint arXiv:2511.03690},
  year = {2025},
  url = {https://arxiv.org/abs/2511.03690}
}

@inproceedings{yang2024sweagent,
  title = {{SWE}-agent: Agent-Computer Interfaces Enable Automated Software Engineering},
  author = {Yang, John and Jimenez, Carlos E. and Wettig, Alexander and Lieret, Kilian and Yao, Shunyu and Narasimhan, Karthik and Press, Ofir},
  booktitle = {Advances in Neural Information Processing Systems},
  year = {2024},
  url = {https://arxiv.org/abs/2405.15793}
}

@inproceedings{xiong2025memorymanagement,
  title = {How Memory Management Impacts {LLM} Agents: An Empirical Study of Experience-Following Behavior},
  author = {Xiong, Guangzhi and Jin, Qiao and Zhang, Zhizheng and Lu, Xiao and Wang, Zhiyong and Ma, Meng and Wang, Xunzhu and Wang, Yiyang and Liu, Yikai and Sun, Huaxiu and Wang, Fei and Liu, Zhiyong and Liu, Chenyan},
  booktitle = {Proceedings of the 2025 Conference on Empirical Methods in Natural Language Processing},
  year = {2025},
  url = {https://arxiv.org/abs/2505.16067}
}

@article{ehtesham2025interoperability,
  title = {A Survey of Agent Interoperability Protocols: Model Context Protocol ({MCP}), Agent Communication Protocol ({ACP}), Agent-to-Agent Protocol ({A2A}), and Agent Network Protocol ({ANP})},
  author = {Ehtesham, Usama and Dib, Salam and Almajali, Sufian and Peixoto, Tiago and Bhattacharya, Jay and Singla, Anupam and Diamantopoulos, Themistoklis},
  journal = {arXiv preprint arXiv:2505.02279},
  year = {2025},
  url = {https://arxiv.org/abs/2505.02279}
}

@inproceedings{wang2026agentspec,
  title = {{AgentSpec}: Customizable Runtime Enforcement for Safe and Reliable {LLM} Agents},
  author = {Wang, Haoyu and Poskitt, Christopher M. and Sun, Jun},
  booktitle = {Proceedings of the 48th IEEE/ACM International Conference on Software Engineering},
  year = {2026},
  url = {https://arxiv.org/abs/2503.18666}
}

@misc{li2026agentharnesssurvey,
  title = {Agent Harness Engineering: A Survey},
  author = {Li, Junjie and Xiao, Xi and Zhang, Yunbei and Liu, Chen and Zhao, Lin and Liao, Xiaoying and Ji, Yingrui and Wang, Janet and Gu, Jianyang and Ge, Yingqiang and Xu, Weijie and Fang, Xi and Xu, Xiang and Zhao, Tianchen and Kim, Youngeun and Wang, Tianyang and Hamm, Jihun and Krishnaswamy, Smita and Huan, Jun and Reddy, Chandan K.},
  year = {2026},
  note = {Under review},
  howpublished = {\url{https://picrew.github.io/LLM-Harness/}}
}

@misc{openai2026harness,
  title = {Harness Engineering: Leveraging {Codex} in an Agent-First World},
  author = {{OpenAI}},
  year = {2026},
  howpublished = {\url{https://openai.com/index/harness-engineering/}}
}

@misc{anthropic2026harness,
  title = {Harness Design for Long-Running Application Development},
  author = {{Anthropic}},
  year = {2026},
  howpublished = {\url{https://www.anthropic.com/engineering/harness-design-long-running-apps}}
}

@misc{langchain2025harnesses,
  title = {Agent Frameworks, Runtimes, and Harnesses - Oh My!},
  author = {{LangChain}},
  year = {2025},
  howpublished = {\url{https://www.langchain.com/blog/agent-frameworks-runtimes-and-harnesses-oh-my}}
}

@misc{langchain2026middleware,
  title = {How Middleware Lets You Customize Your Agent Harness},
  author = {{LangChain}},
  year = {2026},
  howpublished = {\url{https://blog.langchain.com/how-middleware-lets-you-customize-your-agent-harness/}}
}

@article{pan2026nlah,
  title = {Natural-Language Agent Harnesses},
  author = {Pan, Linyue and Zou, Lexiao and Guo, Shuo and Ni, Jingchen and Zheng, Hai-Tao},
  journal = {arXiv preprint arXiv:2603.25723},
  year = {2026},
  url = {https://arxiv.org/abs/2603.25723}
}

@inproceedings{zhang2026ace,
  title = {Agentic Context Engineering: Evolving Contexts for Self-Improving Language Models},
  author = {Zhang, Qizheng and Hu, Changran and Upasani, Shubhangi and Ma, Boyuan and Hong, Fenglu and Kamanuru, Vamsidhar and Rainton, Jay and Wu, Chen and Ji, Mengmeng and Li, Hanchen and Thakker, Urmish and Zou, James and Olukotun, Kunle},
  booktitle = {International Conference on Learning Representations},
  year = {2026},
  url = {https://arxiv.org/abs/2510.04618}
}

@inproceedings{agrawal2026gepa,
  title = {{GEPA}: Reflective Prompt Evolution Can Outperform Reinforcement Learning},
  author = {Agrawal, Lakshya A. and Tan, Shangyin and Soylu, Dilara and Ziems, Noah and Khare, Rishi and Opsahl-Ong, Krista and Singhvi, Arnav and Shandilya, Herumb and Ryan, Michael J. and Jiang, Meng and Potts, Christopher and Sen, Koushik and Dimakis, Alexandros G. and Stoica, Ion and Klein, Dan and Zaharia, Matei and Khattab, Omar},
  booktitle = {International Conference on Learning Representations},
  year = {2026},
  url = {https://arxiv.org/abs/2507.19457}
}

@article{youtu2025tfgrpo,
  title = {Training-Free Group Relative Policy Optimization},
  author = {Cai, Yuzheng and Cai, Siqi and Shi, Yuchen and Xu, Zihan and Chen, Lichao and Qin, Yulei and Tan, Xiaoyu and Li, Gang and Li, Zongyi and Lin, Haojia and Mao, Yong and Li, Ke and Sun, Xing},
  journal = {arXiv preprint arXiv:2510.08191},
  year = {2025},
  url = {https://arxiv.org/abs/2510.08191}
}

@article{lin2026ahe,
  title = {Agentic Harness Engineering: Observability-Driven Automatic Evolution of Coding-Agent Harnesses},
  author = {Lin, Jiahang and Liu, Shichun and Pan, Chengjun and Lin, Lizhi and Dou, Shihan and Xi, Zhiheng and Huang, Xuanjing and Yan, Hang and Han, Zhenhua and Gui, Tao and Jiang, Yu-Gang},
  journal = {arXiv preprint arXiv:2604.25850},
  year = {2026},
  url = {https://arxiv.org/abs/2604.25850}
}

@article{robins1995catastrophic,
  title = {Catastrophic Forgetting, Rehearsal and Pseudorehearsal},
  author = {Robins, Anthony},
  journal = {Connection Science},
  volume = {7},
  number = {2},
  pages = {123--146},
  year = {1995},
  doi = {10.1080/09540099550039318}
}

@article{kirkpatrick2017ewc,
  title = {Overcoming Catastrophic Forgetting in Neural Networks},
  author = {Kirkpatrick, James and Pascanu, Razvan and Rabinowitz, Neil and Veness, Joel and Desjardins, Guillaume and Rusu, Andrei A. and Milan, Kieran and Quan, John and Ramalho, Tiago and Grabska-Barwinska, Agnieszka and Hassabis, Demis and Clopath, Claudia and Kumaran, Dharshan and Hadsell, Raia},
  journal = {Proceedings of the National Academy of Sciences},
  volume = {114},
  number = {13},
  pages = {3521--3526},
  year = {2017},
  doi = {10.1073/pnas.1611835114}
}

@inproceedings{lopezpaz2017gem,
  title = {Gradient Episodic Memory for Continual Learning},
  author = {Lopez-Paz, David and Ranzato, Marc'Aurelio},
  booktitle = {Advances in Neural Information Processing Systems},
  year = {2017},
  url = {https://arxiv.org/abs/1706.08840}
}

@article{chaudhry2019tiny,
  title = {On Tiny Episodic Memories in Continual Learning},
  author = {Chaudhry, Arslan and Rohrbach, Marcus and Elhoseiny, Mohamed and Ajanthan, Thalaiyasingam and Dokania, Puneet K. and Torr, Philip H. S. and Ranzato, Marc'Aurelio},
  journal = {arXiv preprint arXiv:1902.10486},
  year = {2019},
  url = {https://arxiv.org/abs/1902.10486}
}

@article{vandeven2019three,
  title = {Three Scenarios for Continual Learning},
  author = {van de Ven, Gido M. and Tolias, Andreas S.},
  journal = {arXiv preprint arXiv:1904.07734},
  year = {2019},
  url = {https://arxiv.org/abs/1904.07734}
}

@article{fang2025selfevolving,
  title = {A Comprehensive Survey of Self-Evolving {AI} Agents: A New Paradigm Bridging Foundation Models and Lifelong Agentic Systems},
  author = {Fang, Jinyuan and Peng, Yanwen and Zhang, Xi and Wang, Yingxu and Yi, Xinhao and Zhang, Guibin and Xu, Yi and Wu, Bin and Liu, Siwei and Li, Zihao and Ren, Zhaochun and Aletras, Nikos and Wang, Xi and Zhou, Han and Meng, Zaiqiao},
  journal = {arXiv preprint arXiv:2508.07407},
  year = {2025},
  url = {https://arxiv.org/abs/2508.07407}
}

@inproceedings{yuan2024selfrewarding,
  title = {Self-Rewarding Language Models},
  author = {Yuan, Weizhe and Pang, Richard Yuanzhe and Cho, Kyunghyun and Li, Xian and Sukhbaatar, Sainbayar and Xu, Jing and Weston, Jason},
  booktitle = {International Conference on Machine Learning},
  year = {2025},
  url = {https://arxiv.org/abs/2401.10020}
}

@article{kumar2024score,
  title = {Training Language Models to Self-Correct via Reinforcement Learning},
  author = {Kumar, Aviral and Zhuang, Vincent and Agarwal, Rishabh and Su, Yi and Co-Reyes, John D. and Singh, Avi and Baumli, Kate and Iqbal, Shariq and Bishop, Colton and Roelofs, Rebecca and Zhang, Lei M. and McKinney, Kay and Shrivastava, Disha and Paduraru, Cosmin and Tucker, George and Precup, Doina and Behbahani, Feryal and Faust, Aleksandra},
  journal = {arXiv preprint arXiv:2409.12917},
  year = {2024},
  url = {https://arxiv.org/abs/2409.12917}
}

@inproceedings{setlur2025rewarding,
  title = {Rewarding Progress: Scaling Automated Process Verifiers for {LLM} Reasoning},
  author = {Setlur, Amrith and Nagpal, Chirag and Fisch, Adam and Geng, Xinyang and Eisenstein, Jacob and Agarwal, Rishabh and Agarwal, Alekh and Berant, Jonathan and Kumar, Aviral},
  booktitle = {International Conference on Learning Representations},
  year = {2025},
  url = {https://arxiv.org/abs/2410.08146}
}

@article{choudhury2025agentprm,
  title = {Process Reward Models for {LLM} Agents: Practical Framework and Directions},
  author = {Choudhury, Sanjiban},
  journal = {arXiv preprint arXiv:2502.10325},
  year = {2025},
  url = {https://arxiv.org/abs/2502.10325}
}

@article{wang2025ragen,
  title = {{RAGEN}: Understanding Self-Evolution in {LLM} Agents via Multi-Turn Reinforcement Learning},
  author = {Wang, Zihan and Wang, Kangrui and Wang, Qineng and Zhang, Pingyue and Li, Linjie and Yang, Zhengyuan and Jin, Xing and Yu, Kefan and Nguyen, Minh Nhat and Liu, Licheng and Gottlieb, Eli and Lu, Yiping and Cho, Kyunghyun and Wu, Jiajun and Fei-Fei, Li and Wang, Lijuan and Choi, Yejin and Li, Manling},
  journal = {arXiv preprint arXiv:2504.20073},
  year = {2025},
  url = {https://arxiv.org/abs/2504.20073}
}

@article{feng2025retool,
  title = {{ReTool}: Reinforcement Learning for Strategic Tool Use in {LLM}s},
  author = {Feng, Jiazhan and Huang, Shijue and Qu, Xingwei and Zhang, Ge and Qin, Yujia and Zhong, Baoquan and Jiang, Chengquan and Chi, Jinxin and Zhong, Wanjun},
  journal = {arXiv preprint arXiv:2504.11536},
  year = {2025},
  url = {https://arxiv.org/abs/2504.11536}
}

@inproceedings{zhuge2024gptswarm,
  title = {Language Agents as Optimizable Graphs},
  author = {Zhuge, Mingchen and Wang, Wenyi and Kirsch, Louis and Faccio, Francesco and Khizbullin, Dmitrii and Schmidhuber, J{\"u}rgen},
  booktitle = {International Conference on Machine Learning},
  year = {2024},
  url = {https://arxiv.org/abs/2402.16823}
}

@article{zhang2025aflow,
  title = {{AFlow}: Automating Agentic Workflow Generation},
  author = {Zhang, Jiayi and Xiang, Jinyu and Yu, Zhaoyang and Teng, Fengwei and Chen, Xionghui and Chen, Jiaqi and Zhuge, Mingchen and Cheng, Xin and Hong, Sirui and Wang, Jinlin and Zheng, Bingnan and Liu, Bang and Luo, Yuyu and Wu, Chenglin},
  journal = {arXiv preprint arXiv:2410.10762},
  year = {2025},
  url = {https://arxiv.org/abs/2410.10762}
}

@article{wang2024awm,
  title = {Agent Workflow Memory},
  author = {Wang, Zora Zhiruo and Mao, Jiayuan and Fried, Daniel and Neubig, Graham},
  journal = {arXiv preprint arXiv:2409.07429},
  year = {2024},
  url = {https://arxiv.org/abs/2409.07429}
}

@article{yuan2025evoagent,
  title = {{EvoAgent}: Towards Automatic Multi-Agent Generation via Evolutionary Algorithms},
  author = {Yuan, Siyu and Song, Kaitao and Chen, Jiangjie and Tan, Xu and Li, Dongsheng and Yang, Deqing},
  journal = {arXiv preprint arXiv:2406.14228},
  year = {2025},
  url = {https://arxiv.org/abs/2406.14228}
}

@article{xu2025amem,
  title = {{A-MEM}: Agentic Memory for {LLM} Agents},
  author = {Xu, Wujiang and Liang, Zujie and Mei, Kai and Gao, Hang and Tan, Juntao and Zhang, Yongfeng},
  journal = {arXiv preprint arXiv:2502.12110},
  year = {2025},
  url = {https://arxiv.org/abs/2502.12110}
}

@article{tang2025agentkb,
  title = {Agent {KB}: Leveraging Cross-Domain Experience for Agentic Problem Solving},
  author = {Tang, Xiangru and Qin, Tianrui and Peng, Tianhao and Zhou, Ziyang and Shao, Daniel and Du, Tingting and Wei, Xinming and Xia, Peng and Wu, Fang and Zhu, He and Zhang, Ge and Liu, Jiaheng and Wang, Xingyao and Hong, Sirui and Wu, Chenglin and Cheng, Hao and Wang, Chi and Zhou, Wangchunshu},
  journal = {arXiv preprint arXiv:2507.06229},
  year = {2025},
  url = {https://arxiv.org/abs/2507.06229}
}

@article{yehudai2026agenteval,
  title = {Survey on Evaluation of {LLM}-based Agents},
  author = {Yehudai, Asaf and Eden, Lilach and Li, Alan and Uziel, Guy and Zhao, Yilun and Bar-Haim, Roy and Cohan, Arman and Shmueli-Scheuer, Michal},
  journal = {arXiv preprint arXiv:2503.16416},
  year = {2026},
  url = {https://arxiv.org/abs/2503.16416}
}

\appendix

\FloatBarrier
\clearpage
\raggedbottom

\section{Additional Method Details}
\label{app:protocol-details}

This appendix follows the structure of the main text.
Appendix~\ref{app:protocol-details} expands the method and protocol details from Section~\ref{sec:framework}.
Appendix~\ref{app:experimental-details} describes the benchmarks, baselines, concrete settings, and recorded compute budget used by Section~\ref{sec:experiments}.
Appendix~\ref{app:additional-results} contains supplementary learning curves and replay diagnostics.
Appendix~\ref{app:integration-details} and Appendix~\ref{app:metric-details} summarize implementation and metric records needed to inspect and recompute the results.

\subsection{Evolution Schedule Fields}

The experiments use an epoch-based batch setting, but the same protocol can instantiate other self-evolution schedules.
The experiment configuration records the following schedule fields:
\begin{itemize}
  \item \textbf{state persistence}: whether the harness state is reset between task episodes or persists across them;
  \item \textbf{task reuse}: whether the update-bearing task stream is one pass or repeated across epochs;
  \item \textbf{train size}: the number of tasks selected from the source train pool for each epoch;
  \item \textbf{batch size}: the number of task episodes run before each train-batch update;
  \item \textbf{number of epochs}: the number of traversals over the selected train set;
  \item \textbf{updates per batch}: the number of rollout-update attempts on each train batch;
  \item \textbf{assessment timing}: the boundaries at which frozen snapshots are evaluated.
\end{itemize}
In the default setting, state is persistent, tasks are reused across epochs, updates occur after train batches, and frozen update-validation assessment occurs at epoch end.

\begin{algorithm}[t]
\caption{\textsc{SEAGym} Epoch/Batch Evaluation}
\label{alg:seagym}
\begin{algorithmic}[1]
\Require task index, train/val/test splits, schedule, seed
\Require self-evolving baseline with rollout policy and update rule $U$
\State Sample train batches $\{B_t\}$ and select $V_{\text{update-val}}$, replay views, and final views
\State Record initial snapshot $A_0=(M,H_0)$ as $E_0$
\For{epoch $=1,\ldots,N$}
  \For{train batch $B_t$}
    \State Run task episodes from $B_t$ and collect $(\mathcal{T}_t,F_t)$
    \State Update harness state: $H_{t+1}\leftarrow U(H_t,B_t,\mathcal{T}_t,F_t)$
    \State Save update summary and optional checkpoint
  \EndFor
  \State Freeze current agent as $E_i$ and evaluate on $V_{\text{update-val}}$ \Comment{epoch-end assessment}
\EndFor
\State Evaluate $A_0$ and $A_T$ on final ID, OOD, and replay views \Comment{held-out assessment}
\end{algorithmic}
\end{algorithm}

\subsection{Saved Evaluation Views}

Base manifests contain only train, validation, and test task ids.
Before a scored run, the data module selects and saves:
\begin{itemize}
  \item train batches from the train split;
  \item $V_{\text{update-val}}$ from the validation split;
  \item final ID transfer views from held-out source test tasks;
  \item final OOD transfer views from held-out target-domain test tasks;
  \item replay or diagnostic views when enabled.
\end{itemize}
The saved task ids are part of the run artifact so that metric computation can be recomputed without relying on in-memory sampling state.

\section{Experimental Details}
\label{app:experimental-details}

\subsection{Benchmarks}

The paper experiments use two Harbor-backed benchmark sources.
Terminal-Bench 2.0 provides executable command-line and software-engineering tasks with containerized environments and verifiers \citep{merrill2026terminalbench}.
HLE provides expert-level question-answering tasks \citep{phan2025hle}; we use text-only Math and Physics tasks as source tasks, and text-only CS/AI plus Engineering tasks as held-out OOD transfer tasks.
Harbor provides the task execution substrate, including environments, parallel jobs, trial artifacts, and verifier result files \citep{Harbor_Framework}.
\textsc{SEAGym} does not copy benchmark definitions; it stores task ids, stable attributes, split membership, schedule records, snapshot records, and normalized metric inputs.

\subsection{Baselines}
\label{app:baselines}

We connect self-evolving methods through the rollout/update interface described in Section~\ref{sec:framework}.
ACE evolves persistent context from task traces \citep{zhang2026ace}.
TF-GRPO uses grouped rollout evidence to update an experience/context store without model-weight training \citep{youtu2025tfgrpo}.
AHE edits a broader agent harness, including prompts, middleware, memory, and project files, using observability from previous rollouts \citep{lin2026ahe}.
For all methods, task execution remains Harbor-backed and the method wrapper preserves native update semantics as much as possible.

\subsection{Paper Experiment Settings}

Table~\ref{tab:paper-experiment-settings} lists the full setting used by the main experiments and ablations.

\begin{table*}[t]
\centering
\small
\begin{tabular}{@{}p{0.22\linewidth}p{0.70\linewidth}@{}}
\toprule
\textbf{Setting} & \textbf{Value}\\
\midrule
Task source &
Terminal-Bench 2.0 plus HLE text-only Math and Physics for source training, update-validation, and in-domain testing.\\
Source split sizes &
80 train tasks, 35 validation tasks, and 55 in-domain test tasks.\\
OOD transfer test source &
HLE text-only CS/AI and Engineering, used only as held-out OOD transfer tests.\\
OOD transfer test size &
80 total tasks.\\
Train feedback &
Train tasks provide task trajectories, public errors, verifier rewards, and method-visible artifacts according to the task interface.\\
Validation feedback &
$V_{\text{update-val}}$ is frozen before scored runs and is used only for epoch-end assessment.\\
Final test feedback &
ID and OOD transfer tests are run only from frozen snapshots and are not fed back to the update rule.\\
State behavior &
Agent harness state persists across train batches and epochs. Final evaluation compares the initial snapshot $A_0$ with the final snapshot $A_T$.\\
Main schedule &
Five epochs, train batch size 20, and epoch-end frozen update-validation.\\
Update attempts &
All main-table runs use one SEAGym update attempt per train batch. Method-native rollouts, such as TF-GRPO train-only grouped evidence, are recorded separately from SEAGym update attempts.\\
Rollout attempts &
Validation and final views use single-attempt evaluation. TF-GRPO preserves train-only grouped rollout evidence for its native update rule.\\
Main methods &
ACE, TF-GRPO, and AHE. Each method uses its native harness update rule and saved update artifacts.\\
Main model backend &
DeepSeek-V4-Flash for the main method comparison; exact provider metadata is recorded with the run artifacts.\\
Batch-size ablation &
AHE with batch sizes 10, 40, and 80, compared against the main batch size 20 schedule.\\
Source-choice ablation &
AHE on HLE-only Math and Physics source tasks, compared with the combined Terminal-Bench plus HLE source setting.\\
Cross-model ablation &
AHE with GLM-5.1 and GPT-5.4, compared against the DeepSeek-V4-Flash AHE run.\\
Execution backend &
Harbor-backed task execution with SEAGym snapshot, schedule, metric, and artifact records.\\
Runtime budget &
Scored paper configs use task timeout 1800 seconds and concurrency 16 unless overridden by a run-specific config.\\
Primary metric &
Task success rate on frozen validation and test views, reported as a percentage.\\
Derived metrics &
Update-validation gain, final in-domain gain, OOD gain, forgetting, cost, and token usage when available; gain values are percentage-point differences.\\
Diagnostics &
Update status, setup failures, verifier failures, provider errors, runtime corruption, replay or regression probes when enabled, and harness-state diffs when available.\\
\bottomrule
\end{tabular}
\caption{Full experiment setting for the main runs and ablations. Run-specific overrides are recorded in the corresponding config and artifact files.}
\label{tab:paper-experiment-settings}
\end{table*}

\subsection{Experiment-Specific Configurations}

Table~\ref{tab:experiment-specific-configs} lists the concrete run configuration behind each experiment subsection.

\begin{table*}[t]
\centering
\scriptsize
\setlength{\tabcolsep}{3.2pt}
\begin{tabular}{@{}p{0.16\textwidth}p{0.18\textwidth}p{0.19\textwidth}p{0.13\textwidth}p{0.25\textwidth}@{}}
\toprule
\textbf{Paper section} & \textbf{Training runs} & \textbf{Source / schedule} & \textbf{Model backend} & \textbf{Extra evaluations and reported view}\\
\midrule
Baseline Results &
AHE main, ACE main, and TF-GRPO main &
Terminal-Bench + HLE; 80/35/55 split; five epochs; batch size 20; one update per train batch &
DeepSeek v4 Flash for rollout and update where applicable &
All methods report ID and OOD evaluations from their best-validation snapshots; for AHE, the best-validation snapshot is the final snapshot. The table reports validation, ID, OOD, token cost, and update status.\\
\addlinespace
Training Fix and Forgetting &
AHE main &
Same as AHE main; fixed saved snapshots at initialization, epochs 1--4, and the final epoch &
DeepSeek v4 Flash &
All fixed snapshots are evaluated on the 80 source train tasks. The grid reports success, failure, and rollout-error status by original train exposure order.\\
\addlinespace
Effect of Batch Size &
AHE batch-size sweep: 10, 20, 40, and 80 &
Terminal-Bench + HLE; five epochs; batch sizes 10/20/40/80; one update per train batch &
DeepSeek v4 Flash &
Final ID evaluations use each run's final snapshot. We report this ablation on validation and ID views; OOD is reserved for the main and OOD analyses.\\
\addlinespace
Source Diversity &
Mixed-source AHE and HLE-only AHE &
Main source: Terminal-Bench + HLE 80/35/55. HLE-only source: HLE Math/Physics 80/35/50. Both use five epochs and batch size 20 &
DeepSeek v4 Flash &
Main source uses final ID and OOD evaluations. HLE-only reports final validation/ID and the best-validation OOD snapshot because the final HLE-only snapshot collapses.\\
\addlinespace
Cross-Model Transfer &
AHE evolved with DeepSeek-V4-Flash, GLM-5.1, and GPT-5.4 &
Terminal-Bench + HLE; five epochs; batch size 20; one update per train batch &
DeepSeek-V4-Flash, GLM-5.1, and GPT-5.4, one model per training run &
Cross-backend snapshot evaluations swap evolved AHE snapshots and rollout backends on ID and OOD views. The table reports finalized ID and OOD evaluation records and treats runtime-corrupted trials as diagnostic artifacts rather than scored improvements.\\
\bottomrule
\end{tabular}
\caption{Experiment-specific configurations for the results in Section~\ref{sec:experiments}. Each run artifact records the resolved JSON config, split, model, backend, and snapshot metadata.}
\label{tab:experiment-specific-configs}
\end{table*}

\subsection{Training Runs, Tokens, and Recorded Runtime}

The compact records below summarize the training runs used for the reported results.
Token counts are read from normalized SEAGym metric records.
Rollout tokens count task-execution records, and update tokens count SEAGym baseline-update records when the native method exposes token usage.
Runtime is the sum of unique Harbor-reported task-job runtimes recovered from saved task results.
It is therefore a recorded task-execution runtime, not a complete end-to-end wall-clock measurement; native update wall time and external queueing delays were not recorded consistently across methods.

\begin{center}
\centering
\tiny
\renewcommand{\arraystretch}{1.06}
\textbf{Main runs}\\[2pt]
\begin{tabular}{@{}p{0.20\linewidth}p{0.18\linewidth}p{0.23\linewidth}p{0.25\linewidth}@{}}
\toprule
\textbf{Run} & \textbf{Schedule / rows} & \textbf{Recorded budget} & \textbf{Use}\\
\midrule
Main AHE & b20/e5/u20; 720 rows & 1053.8M rollout; 78.1M update; 14h06m & Main result and diagnostics\\
Main ACE & b20/e5/u20; 720 rows & 1384.0M rollout; update --; 13h55m & Main result\\
Main TF-GRPO & b20/e5/u20; 1120 rows & 2615.1M rollout; 32.0M update; time -- & Main result; grouped rollout evidence\\
\bottomrule
\end{tabular}
\end{center}

\begin{center}
\centering
\tiny
\renewcommand{\arraystretch}{1.06}
\textbf{Ablation and cross-model runs}\\[2pt]
\begin{tabular}{@{}p{0.20\linewidth}p{0.18\linewidth}p{0.23\linewidth}p{0.25\linewidth}@{}}
\toprule
\textbf{Run} & \textbf{Schedule / rows} & \textbf{Recorded budget} & \textbf{Use}\\
\midrule
AHE batch 10 & b10/e5/u40; 720 rows & 567.8M rollout; 125.2M update; 17h47m & Batch-size ablation\\
AHE batch 40 & b40/e5/u10; 720 rows & 782.2M rollout; 33.6M update; 10h44m & Batch-size ablation\\
AHE batch 80 & b80/e5/u5; 720 rows & 769.9M rollout; 17.9M update; 10h09m & Batch-size ablation\\
AHE batch 80 cont. & b80/e5/u5; 720 rows & 896.2M rollout; 16.9M update; 7h40m & Diagnostic continuation\\
AHE HLE only & b20/e5/u20; 710 rows & 625.4M rollout; 68.9M update; time -- & Source-diversity ablation\\
AHE GLM & b20/e5/u20; 720 rows & 159.8M rollout; 111.0M update; 14h32m & Cross-model diagnostic\\
\bottomrule
\end{tabular}
\end{center}

Schedule shorthand reports batch size, epochs, and updates. Task rows include train, update-validation, and final task executions saved by the training run. Additional evaluate-only snapshot jobs are reported separately in their own artifacts and are not added here. A dash means the field was not recoverable from normalized records.

\section{Evolution Artifacts and Case Studies}
\label{app:evolution-artifacts-case-studies}

\subsection{Baseline Evolution Case Study}
\label{app:baseline-evolution-case-study}

The main baseline table reports the performance trajectory of ACE, TF-GRPO, and AHE.
To understand what changed during training, we inspect the saved update artifacts, agent snapshots, and representative task traces.
The three methods do not update the same kind of state.
ACE mainly stores process-level skills, TF-GRPO stores task-family experiences, and AHE edits the runnable harness itself.
Table~\ref{tab:baseline-update-artifacts} summarizes these artifacts before we discuss individual cases.

\begin{table*}[t]
\centering
\small
\setlength{\tabcolsep}{4.5pt}
\begin{tabular}{@{}p{0.14\textwidth}p{0.23\textwidth}p{0.37\textwidth}p{0.18\textwidth}@{}}
\toprule
\textbf{Method} & \textbf{Evolution artifact} & \textbf{Representative evidence} & \textbf{Primary update target}\\
\midrule
ACE &
Skillbook entries &
The selected E4 snapshot contains 13 active skills covering output readback, format checking, explicit constraint checking, dependency installation, clean-build recovery, and writing a candidate answer before time exhaustion. &
Prompt-visible procedural reminders\\
\addlinespace
TF-GRPO &
Experience memory &
The experience store records task-family strategies such as running failing tests first, using QMP sockets for QEMU input, performing round-trip checks for data pipelines, using unsigned reads in binary parsers, and inspecting model label maps before inference. &
Task-family strategies\\
\addlinespace
AHE &
Harness, tool, and middleware patches &
The update manifests add file, web, and session tools; context compaction; HLE completion enforcement; and message-content contract alignment for middleware guidance. &
Runtime behavior and completion protocol\\
\bottomrule
\end{tabular}
\caption{Saved evolution artifacts for the baseline methods. The methods improve different parts of the agent state, so their gains and failure modes are not directly interchangeable.}
\label{tab:baseline-update-artifacts}
\end{table*}

\paragraph{ACE.}
ACE converts past trajectories into a persistent skillbook.
In the selected E4 snapshot, the active skills are mostly transferable execution habits rather than task-specific programs.
For example, \texttt{context-00001} instructs the agent to read back output files and check exact formatting after writing them; \texttt{context-00002} asks the agent to enumerate every explicit problem constraint before finalizing; \texttt{context-00005} pushes the agent to write a candidate answer before spending the remaining budget on repeated derivation; and \texttt{harness-00010} recommends installing a missing tool or package before abandoning an approach.
These artifacts explain why ACE can yield positive gains: it has learned reusable process knowledge from earlier rollouts.

\begin{table*}[t]
\centering
\small
\setlength{\tabcolsep}{4.2pt}
\begin{tabular}{@{}p{0.15\textwidth}p{0.48\textwidth}p{0.27\textwidth}@{}}
\toprule
\textbf{Skill ID} & \textbf{Content summary} & \textbf{Intended behavioral change}\\
\midrule
\texttt{context-00001} &
After writing an output file, immediately read it back and check the exact required format, including absence of Markdown, extra spaces, wrong delimiters, or missing fields. &
Reduce output-format and parsing failures.\\
\addlinespace
\texttt{context-00002} &
After deriving a candidate answer, enumerate the explicit constraints in the task statement and independently verify each one. &
Reduce cases where partial correctness is mistaken for complete correctness.\\
\addlinespace
\texttt{context-00005} &
When reasoning is complete or time is running out, write a candidate answer first instead of repeatedly deriving without producing output. &
Reduce timeout and no-output failures.\\
\addlinespace
\texttt{harness-00008} &
After a multi-target build is interrupted, prefer a clean build rather than piecemeal recovery from partial state. &
Reduce verifier failures caused by stale artifacts or partial builds.\\
\addlinespace
\texttt{harness-00010} &
When a command or dependency is missing, first try to install or provision it instead of immediately abandoning or bypassing the approach. &
Reduce avoidable failures caused by missing tools.\\
\bottomrule
\end{tabular}
\caption{Representative ACE skills that support successful behavior changes. These entries are prompt-visible procedural reminders: they can improve execution hygiene, but they do not themselves enforce verifier-equivalent constraints.}
\label{tab:ace-successful-skills}
\end{table*}

The failed traces in Table~\ref{tab:ace-failure-cases}, however, illustrate one reason the gain remains limited.
The skills are visible to the model, but they do not change the tools, the completion protocol, or the verifier-equivalent checks available during rollout.

\begin{table*}[t]
\centering
\small
\setlength{\tabcolsep}{4.2pt}
\begin{tabular}{@{}p{0.17\textwidth}p{0.31\textwidth}p{0.26\textwidth}p{0.18\textwidth}@{}}
\toprule
\textbf{Task} & \textbf{Observed behavior} & \textbf{Verifier failure} & \textbf{Mechanistic implication}\\
\midrule
\texttt{polyglot-rust-c} &
The agent wrote \texttt{main.rs}, compiled it with both \texttt{rustc} and \texttt{g++}, and tested that the two executables produced matching Fibonacci outputs. &
The final directory still contained \texttt{cmain} and \texttt{main}; the verifier expected \texttt{/app/polyglot} to contain only \texttt{main.rs}. &
Local functional testing did not enforce the final submission-state constraint.\\
\addlinespace
\texttt{video-processing} &
The script ran, wrote \texttt{output.toml}, and produced parseable fields. &
The predicted landing frames were 61 and 229, while the verifier accepted only [62, 64] and [231, 234]. &
Generic output checks did not create a task-specific temporal oracle.\\
\addlinespace
\texttt{db-wal-recovery} &
The recovered JSON existed, parsed correctly, had the expected structure, and satisfied ordering checks. &
WAL semantics were incomplete: for example, record \texttt{id=1} kept value 100 instead of the expected updated value 150. &
Structural validity did not imply recovery of the core state-update semantics.\\
\addlinespace
\texttt{path-tracing-reverse} &
The skillbook included a dependency-installation lesson, but the rollout still encountered a missing \texttt{file} command. The task also prohibited invoking the original binary, yet the rollout ran \texttt{./mystery}. &
The verifier failed after the generated reverse program did not satisfy the isolated execution check. &
Passive skills do not guarantee that critical constraints are applied at the right decision point.\\
\bottomrule
\end{tabular}
\caption{Representative ACE failure cases. The traces support a narrow interpretation: ACE improves execution hygiene, but prompt-visible skills do not reliably enforce final-state constraints, task-specific oracles, or critical prohibitions.}
\label{tab:ace-failure-cases}
\end{table*}

The \texttt{polyglot-rust-c} trace is especially informative because the agent did perform meaningful local validation.
It confirmed that the Rust and C++ executables compiled and matched on several inputs.
The failure occurred after this local verification step: temporary build artifacts were left in the submission directory.
This trace is consistent with ACE encouraging checks that an artifact works, while still missing constraints about the final filesystem state.
Similarly, \texttt{video-processing} and \texttt{db-wal-recovery} show that output existence, parseability, and simple structure checks are not enough when the success condition depends on fine-grained semantic alignment.
The evidence therefore supports a bounded conclusion: ACE transfers useful procedural knowledge, but the skills often stop short of the verifier condition.

\paragraph{TF-GRPO.}
TF-GRPO stores a larger experience memory.
Compared with ACE skills, these entries are often more task-family specific.
Table~\ref{tab:tfgrpo-experience-cases} gives examples from the saved experience store.

\begin{table*}[t]
\centering
\small
\setlength{\tabcolsep}{4.2pt}
\begin{tabular}{@{}p{0.12\textwidth}p{0.47\textwidth}p{0.31\textwidth}@{}}
\toprule
\textbf{Experience} & \textbf{Stored lesson} & \textbf{Task-family relevance}\\
\midrule
\texttt{G0} &
For unfamiliar codebases and vulnerability fixes, run the existing test suite first and use failing tests to identify the root cause. &
Code repair and vulnerability-fixing tasks.\\
\addlinespace
\texttt{G1}, \texttt{G79} &
For programmatic QEMU input, use QMP over a TCP socket and prefer \texttt{send-key}; verify the channel with a small Python socket script. &
VM and emulator-interaction tasks.\\
\addlinespace
\texttt{G101} &
After validating structural constraints in data pipelines, run forward-plus-inverse round-trip checks and confirm clean-environment determinism with no leftover artifacts. &
Data workflow and pipeline-verification tasks.\\
\addlinespace
\texttt{G102}, \texttt{G99} &
Use unsigned integer reads for binary data and respect segment sizes when extracting ELF/PE data. &
Binary parsing and systems tasks.\\
\addlinespace
\texttt{G109}, \texttt{G131} &
When reconstructing procedurally generated images, compare key pixels or extract exact constants before full rendering. &
Reverse-engineering and generated-image tasks.\\
\addlinespace
\texttt{G113}, \texttt{G74} &
Before serving a pretrained Hugging Face classifier, inspect the model configuration's \texttt{id2label} mapping. &
Model inference and deployment tasks.\\
\bottomrule
\end{tabular}
\caption{Representative TF-GRPO experiences. TF-GRPO records more concrete task-family strategies than ACE's general process reminders.}
\label{tab:tfgrpo-experience-cases}
\end{table*}

These experiences help explain why TF-GRPO can produce a large source-validation gain.
They do not merely say ``verify more''; they encode specific actions for recurring task types, such as using QMP for QEMU interaction, checking round trips in data pipelines, or avoiding signed overflow in binary parsing.
Several successful TF-GRPO evaluations occur on task families compatible with these lessons, including \texttt{pytorch-model-recovery}, \texttt{build-pmars}, \texttt{portfolio-optimization}, \texttt{llm-inference-batching-scheduler}, \texttt{kv-store-grpc}, \texttt{compile-compcert}, \texttt{vulnerable-secret}, \texttt{fix-code-vulnerability}, \texttt{hf-model-inference}, and \texttt{reshard-c4-data}.
We do not attribute any single success to a single memory entry, because a rollout may combine multiple experiences and model choices.
The evidence does support the more conservative claim that TF-GRPO updates a more concrete library of task-family priors than ACE.

This also clarifies the boundary of TF-GRPO.
Its experiences can make a later attempt more targeted when a similar situation arises, but they still operate through the model's prompt-time behavior.
They do not add tools, intercept completion, or enforce a runtime check.
Thus, TF-GRPO sits between ACE and AHE: it is more task-specific than a short skillbook, but it still depends on retrieval and adoption during rollout.

\paragraph{AHE.}
AHE produces the most direct changes to the execution path.
Its update manifests record not only what changed, but also the failure pattern targeted by each change and why that component was edited.
Table~\ref{tab:ahe-update-cases} summarizes representative updates.

\begin{table*}[t]
\centering
\small
\setlength{\tabcolsep}{3.8pt}
\begin{tabular}{@{}p{0.10\textwidth}p{0.28\textwidth}p{0.30\textwidth}p{0.24\textwidth}@{}}
\toprule
\textbf{Iter.} & \textbf{Concrete update} & \textbf{Targeted failure pattern} & \textbf{Why this component matters}\\
\midrule
1 &
Added file, web, session, and context-management tools, including direct file operations, web search/read, explicit completion, progress tracking, and context compaction. &
Initial rollouts relied on shell workarounds, answered HLE questions from recall, lacked explicit completion control, and could overflow context on long tasks. &
The harness gains new action channels and a middleware layer that later updates can constrain.\\
\addlinespace
4 &
Added an HLE verification enforcer that intercepts completion unless research and verification evidence are present. &
Prompt instructions asked for research and verification, but rollouts still skipped them. &
The final completion point is programmatically constrained.\\
\addlinespace
9 &
Strengthened HLE verification and added an artifact-cleanup rule for implementation tasks. &
Some HLE answers were ``verified'' by one flawed script, while \texttt{polyglot-c-py} failed after leaving an extra compiled binary in the output directory. &
The update changes what counts as sufficient verification and adds a concrete submission-state rule.\\
\addlinespace
14 &
Repaired tool-error recovery and added middleware checks for heredoc-heavy coding and sleep-chaining patterns. &
\texttt{path-tracing} repeatedly used large inline heredocs, and \texttt{rstan-to-pystan} used long chained sleeps; the recovery middleware itself was also not firing correctly. &
Runtime guidance can intervene during inefficient execution loops rather than relying only on prompt-level advice.\\
\addlinespace
16 &
Added multiple-choice detection and option-specific research requirements to HLE verification. &
For HLE multiple-choice tasks, the agent searched general topics and wrote verification code that confirmed the same hypothesis instead of checking each option. &
Verification becomes tied to the answer format and requires evidence about competing choices.\\
\addlinespace
17 &
Added handling for failed web search, dismissed contradictory evidence, and manual combinatorial reasoning. &
Some HLE rollouts guessed after repeated no-result searches, treated counter-evidence as a special case, or reasoned through counting problems without code. &
The middleware and prompt steer the agent toward fallback computation and explicit investigation of counter-evidence.\\
\addlinespace
20 &
Aligned middleware message content with the runtime message contract so guidance from tool-error recovery, verification reminders, and HLE verification can be injected correctly. &
Middleware guidance could fail to enter the model context when message content did not match the runtime contract. &
The middleware can inject guidance into the model context again.\\
\bottomrule
\end{tabular}
\caption{Representative AHE update artifacts. Unlike ACE and TF-GRPO, AHE often changes the tools, middleware, and completion protocol available to the agent during rollout.}
\label{tab:ahe-update-cases}
\end{table*}

These artifacts help explain why AHE can improve validation, ID, and OOD together in the main result table.
The update does not only remind the model to perform better checks; it can change the tools available for file editing and research, the way long contexts are managed, and the conditions under which a task may be completed.
The successful AHE evaluations include tasks that naturally use these capabilities, such as \texttt{multi-source-data-merger}, \texttt{query-optimize}, \texttt{pytorch-model-recovery}, \texttt{portfolio-optimization}, \texttt{llm-inference-batching-scheduler}, \texttt{vulnerable-secret}, \texttt{fix-git}, \texttt{hf-model-inference}, and \texttt{reshard-c4-data}.
Again, we avoid one-to-one causal attribution between a patch and a task success.
The stronger evidence is that the changed components match the capabilities demanded by many solved tasks: file interaction, iterative verification, tool-mediated research, and explicit completion control.

AHE failures are also informative.
The method still fails on tasks such as \texttt{train-fasttext}, \texttt{make-mips-interpreter}, \texttt{torch-tensor-parallelism}, and some HLE examples, sometimes through timeout, empty model responses, verifier timeout, or remaining task complexity.
Thus, harness-level editing expands what can be changed during a rollout, but it does not remove model-budget limits, long-horizon reasoning failures, or environment instability.

\paragraph{Cross-method interpretation.}
These cases suggest that the three methods improve agents through different mechanisms.
ACE mainly changes what the agent is reminded to do: it can encourage output checking, dependency handling, and local validation, but these reminders may not be applied at the right moment or may stop short of the final verifier condition.
TF-GRPO stores more task-specific experiences, such as using QMP for QEMU interaction or performing round-trip checks for data pipelines, which can make the agent's next attempt more targeted when a similar situation arises.
AHE changes the execution environment more directly by adding tools, middleware, and completion constraints, so its updates can affect not only the agent's reasoning but also the actions available during a rollout.
This distinction helps explain why improvements differ across methods: the learned artifact determines where training can affect the next rollout, and failures often occur just beyond that scope.

\subsection{Training Forgetting Case Study}
\label{app:training-forgetting-case-study}

The train-replay experiment is not a second report of train score.
It checks whether the evolved agent retains capabilities on seen tasks, and whether new capabilities are gained by sacrificing tasks that were already solved.
Because each saved snapshot is replayed on the same 80 source-training tasks, the replay grid lets us inspect how a harness update changes concrete task outcomes.
Table~\ref{tab:train-forgetting-snapshots} gives the snapshot-level evidence before we analyze individual tasks.
In the case-study tables, \texttt{tb/} abbreviates \texttt{terminal-bench/}.

\begin{table*}[t]
\centering
\small
\setlength{\tabcolsep}{4.2pt}
\begin{tabular}{@{}p{0.10\textwidth}rrp{0.58\textwidth}@{}}
\toprule
\textbf{Snapshot} & \textbf{Solved / 80} & \textbf{Rollout errors} & \textbf{Evidence / interpretation}\\
\midrule
$A_0$ & 34 & 4 & Initial harness before training.\\
$E_4$ & 38 & 7 & File tools, web tools, completion control, and verification constraints start to help.\\
$E_8$ & 37 & 8 & The harness keeps some gains, but local task churn appears.\\
$E_{12}$ & 42 & 6 & Best intermediate snapshot before the later execution-path collapse.\\
$E_{16}$ & 6 & 66 & Widespread runtime message-contract violation, not ordinary task forgetting.\\
$A_T$ & 43 & 5 & The message contract is restored, and earlier gains become visible again.\\
\bottomrule
\end{tabular}
\caption{Snapshot-level train replay evidence for the AHE main run. The sharp $E_{16}$ drop is dominated by execution errors, while $A_T$ recovers after the runtime message contract is restored.}
\label{tab:train-forgetting-snapshots}
\end{table*}

\begin{table*}[t]
\centering
\scriptsize
\setlength{\tabcolsep}{3.2pt}
\begin{tabular}{@{}p{0.27\textwidth}ccccccp{0.34\textwidth}@{}}
\toprule
\textbf{Task} & \textbf{$A_0$} & \textbf{$E_4$} & \textbf{$E_8$} & \textbf{$E_{12}$} & \textbf{$E_{16}$} & \textbf{$A_T$} & \textbf{Evidence / meaning}\\
\midrule
\texttt{tb/mailman} & F & S & F & F & ERR & S &
Early harness updates solve an initial failure, but the gain is not stable across snapshots.\\
\addlinespace
\texttt{tb/polyglot-c-py} & F & F & F & S & ERR & S &
Success appears after artifact-cleanup rules are added and remains after the runtime path is restored.\\
\addlinespace
\texttt{tb/rstan-to-pystan} & ERR & ERR & ERR & ERR & ERR & S &
Later execution constraints and runtime repair are needed before the long-running task succeeds.\\
\addlinespace
\texttt{hle\_\_66ecb2...} & F & S & F & S & ERR & S &
HLE verification and search updates produce repeated but finally retained gains.\\
\addlinespace
\texttt{hle\_\_67206a...} & F & F & S & S & ERR & S &
The HLE task becomes stable before $E_{16}$; the $E_{16}$ failure is an execution error rather than an answer error.\\
\bottomrule
\end{tabular}
\caption{Representative train-replay task trajectories. S denotes a solved task, F an unsolved non-error trial, and ERR a rollout error. These cases show reusable gains, but also non-monotonic task behavior.}
\label{tab:train-forgetting-task-trajectories}
\end{table*}

This trajectory shows why training forgetting cannot be judged only by whether the final score is higher than the initial score.
AHE has a positive final net effect, but the $E_{16}$ snapshot exposes a different risk: when evolution can modify the harness and middleware, forgetting may appear as a broken execution path rather than as a model that no longer knows how to solve a task.

The concrete task trajectories in Table~\ref{tab:train-forgetting-task-trajectories} connect the aggregate gain to saved update artifacts.
Early gains come from executable harness changes.
Iteration 1 adds file read/write, search, replacement, directory traversal, web search/read, \texttt{complete\_task}, \texttt{write\_todos}, and context-compaction tools and middleware.
Iteration 4 adds an HLE verification enforcer that requires web research and verification code before HLE or knowledge-task submission.
Iteration 9 further strengthens HLE verification and converts the compiled-artifact residue observed in \texttt{polyglot-c-py} into an artifact-cleanup rule before submission.
These are not single-task memories; they change the action space and completion conditions for later rollouts.
Those changes are reflected in replay tasks that move from failure to success:
\texttt{mailman} is solved by an early snapshot and again by the final snapshot;
\texttt{polyglot-c-py} is solved after an update that explicitly records the extra-binary failure pattern and adds cleanup before submission;
\texttt{rstan-to-pystan} is solved only after later runtime-path changes; and two HLE examples become solvable after verification-oriented updates.
These cases justify a limited mechanism claim: the gains come from executable harness behavior, including tools, middleware, verification checks, cleanup rules, and completion control, rather than from memorizing individual training answers.

\begin{table*}[t]
\centering
\small
\setlength{\tabcolsep}{4pt}
\begin{tabular}{@{}p{0.31\textwidth}cccp{0.43\textwidth}@{}}
\toprule
\textbf{Task} & \textbf{$A_0$} & \textbf{$E_{16}$} & \textbf{$A_T$} & \textbf{Explanation}\\
\midrule
\texttt{tb/compile-compcert} & S & ERR & S &
A previously solved task becomes an execution error at $E_{16}$ and recovers after the final runtime-contract fix.\\
\texttt{tb/configure-git-webserver} & S & ERR & S &
The same solved--error--solved pattern indicates execution-path failure rather than task-specific loss.\\
\texttt{tb/mailman} & F & ERR & S &
The final snapshot both restores the runtime path and solves a task that the initial harness failed.\\
\texttt{tb/rstan-to-pystan} & ERR & ERR & S &
The task remains errored until the final snapshot, so recovery is not a simple rollback to $A_0$.\\
\texttt{hle\_\_67254b...} & S & ERR & F &
The runtime error is removed, but the task is still forgotten at the final snapshot.\\
\bottomrule
\end{tabular}
\caption{Representative evidence for the $E_{16}$ runtime collapse. The pattern is dominated by message-contract failures in the execution path, not by uniformly worse task reasoning.}
\label{tab:train-forgetting-runtime-collapse}
\end{table*}

The $E_{16}$ snapshot has a different signature from normal forgetting.
It solves only 6/80 replay tasks and records 66 rollout errors.
Table~\ref{tab:train-forgetting-runtime-collapse} shows that the affected set includes tasks solved both before and after the collapse, tasks eventually fixed by $A_T$, and one task that remains forgotten even after the runtime error is removed.
The saved update summaries identify the dominant cause as a NexAU message-sequence contract violation.
Middleware-injected messages no longer satisfy the typed-message contract expected by the token counter and message schema.
Iteration 19 first replaces legacy dictionary messages with \texttt{Message(role=Role.SYSTEM, content=...)}, but the content is still a plain string.
Iteration 20 then identifies that content must be block-structured and wraps it as \texttt{[TextBlock(text=...)]}.
After the affected middleware paths are repaired, current-batch success recovers from 1/20 to 10/20, replay errors fall from 66 to 5, and $A_T$ gains 38 tasks relative to $E_{16}$ while losing only one.

We therefore do not interpret $E_{16}$ as ordinary catastrophic forgetting.
It exposes the sensitivity of harness evolution to execution contracts: once middleware message construction is changed incorrectly, the error propagates across many tasks that require middleware guidance.
Snapshot replay separates this process failure from ordinary answer failure; otherwise the final $A_T$ score of 43/80 would hide the large intermediate execution-path collapse.

\begin{table*}[t]
\centering
\scriptsize
\setlength{\tabcolsep}{3.2pt}
\begin{tabular}{@{}p{0.27\textwidth}ccccccp{0.34\textwidth}@{}}
\toprule
\textbf{Task} & \textbf{$A_0$} & \textbf{$E_4$} & \textbf{$E_8$} & \textbf{$E_{12}$} & \textbf{$E_{16}$} & \textbf{$A_T$} & \textbf{Forgetting form}\\
\midrule
\texttt{tb/kv-store-grpc} & S & F & S & F & ERR & F &
Alternates across snapshots, suggesting a changed execution path on the same implementation task.\\
\texttt{tb/reshard-c4-data} & S & S & S & S & ERR & ERR &
Stable initial success is lost through an execution error that remains at $A_T$.\\
\texttt{hle\_\_6720ff...} & S & F & S & F & F & F &
Non-monotonic HLE behavior becomes a final loss independent of the $E_{16}$ error spike.\\
\texttt{hle\_\_67254b...} & S & F & F & F & ERR & F &
The runtime error is repaired, but the final answer behavior does not recover.\\
\bottomrule
\end{tabular}
\caption{Initially solved tasks that are absent from the final AHE snapshot. The final gain is not lossless retention: $A_T$ fixes 13 initial failures but forgets 4 initial successes.}
\label{tab:train-forgetting-forgotten-cases}
\end{table*}

The final replay result should therefore be read as task churn with a positive net effect, not as a rollback to the initial agent.
Relative to $A_0$, $A_T$ fixes 13 initially failed tasks and loses 4 initially solved tasks, yielding a net gain of nine tasks on the replay set.
The fixed cases include \texttt{mailman}, \texttt{polyglot-c-py}, \texttt{rstan-to-pystan}, and multiple HLE tasks.
\texttt{polyglot-c-py} is especially informative: $A_0$, $E_4$, and $E_8$ fail, $E_{12}$ succeeds after the artifact-cleanup rule is introduced, $E_{16}$ is interrupted by the runtime message error, and $A_T$ succeeds again after the runtime path is repaired.
This is evidence that an observed failure can be converted into a reusable submission rule and continue to help after a later execution-path repair.
The forgotten cases in Table~\ref{tab:train-forgetting-forgotten-cases} prevent a stronger claim of lossless harness improvement.
AHE does not expand capability while preserving every old solution.
By modifying tools, middleware, and verification policy, it changes the agent's behavior distribution: for some tasks, a failure mode becomes a reusable constraint; for others, the new constraints and tool-use paths can cause a previously successful solution to be skipped or routed through a longer and more fragile verification chain.

The replay diagnostic therefore does not answer only whether AHE forgets.
It decomposes the process into three observable phenomena: reusable harness improvements, genuine task-level forgetting, and transient runtime failures introduced by execution-system updates.
The saved snapshots and metric records make these phenomena distinguishable: the final gain comes from 13 fixed initial failures, the $E_{16}$ collapse is mainly a runtime message-contract failure, and 4 initial successes remain absent from the final snapshot.

\subsection{Batch-Size Case Study}
\label{app:batch-size-case-study}

The batch-size sweep changes how many trajectories AHE must analyze in one update while keeping the train, update-validation, and ID test sets fixed.
The total train-task exposure is held constant, but the number of update calls changes: batch 10 has 40 updates, batch 20 has 20, batch 40 has 10, and batch 80 has 5.
The aggregate curves show a non-monotonic pattern, but the saved update artifacts reveal why batch size affects more than statistical efficiency.
AHE updates are LLM-driven harness edits with a bounded reading and reasoning budget; larger batches do not automatically give the evolve agent proportionally more analysis capacity.

\begin{table*}[t]
\centering
\scriptsize
\setlength{\tabcolsep}{3.6pt}
\begin{tabular}{@{}p{0.07\textwidth}rrrrrp{0.33\textwidth}@{}}
\toprule
\textbf{Batch} & \textbf{Updates} & \textbf{Validation} & \textbf{ID test} & \textbf{Update tok./update} & \textbf{Update tok./train task} & \textbf{Primary evidence}\\
\midrule
10 & 40 & 37.1$\rightarrow$22.9 & 38.2$\rightarrow$23.6 & 3.13M & 0.31M &
Frequent small-sample updates; local fixes coexist with configuration and runtime drift.\\
20 & 20 & 40.0$\rightarrow$57.1 & 40.0$\rightarrow$49.1 & 3.91M & 0.20M &
Best balance between evidence diversity, trace inspection depth, and recovery opportunity.\\
40 & 10 & 37.1$\rightarrow$40.0 & 41.8$\rightarrow$43.6 & 3.36M & 0.08M &
Small positive gain; shared failures are visible, but heterogeneous task-level analysis is thinner.\\
80 & 5 & 42.9$\rightarrow$25.7 & 41.8$\rightarrow$25.5 & 3.57M & 0.04M &
Evidence overload; salient global patterns dominate long-tail failures and the final state remains runtime-unstable.\\
\bottomrule
\end{tabular}
\caption{Batch-size sweep summary. Token fields use recorded update tokens normalized per update and per source-train task.}
\label{tab:batch-size-case-summary}
\end{table*}

\begin{table*}[t]
\centering
\scriptsize
\setlength{\tabcolsep}{3.2pt}
\begin{tabular}{@{}p{0.08\textwidth}p{0.25\textwidth}p{0.35\textwidth}p{0.24\textwidth}@{}}
\toprule
\textbf{Batch} & \textbf{Update evidence} & \textbf{Harness change / observed failure} & \textbf{Mechanism supported}\\
\midrule
10, iter. 16 &
Current batch has 0/10 successes; 9/10 failures are execution exceptions such as runtime, remote-protocol, or attribute errors, and the only genuine task failure is \texttt{regex-chess}. &
The update analyzes \texttt{regex-chess}: the agent repeatedly rewrites \texttt{generate.py}, stays in local regex-escaping and move-generation debugging, does not use web search, and does not switch strategy. It adds rabbit-hole triggers for repeated file rewrites and repeated failed commands. &
Small batches can trigger reasonable but narrow updates from high-variance evidence.\\
\addlinespace
10, iter. 40 &
All 10 tasks fail before task solving starts. &
\texttt{code\_agent.yaml} passes \texttt{max\_interventions: 1} before the HLE verification middleware accepts that parameter; the previous update changed the config without synchronizing the constructor signature. &
Many frequent edits increase runtime and configuration contract risk.\\
\addlinespace
20, iter. 20 &
A shared message-content failure is visible across the current batch; current-batch success recovers from 1/20 to 10/20 after repair. &
Tool-error recovery, verification reminders, and HLE verification were injecting guidance with message content that did not match the NexAU typed-message schema. The update changes those paths to construct system messages as \texttt{Message(..., content=[TextBlock(...)])}; train replay errors later fall from 66 to 5. &
Batch 20 exposes cross-task failures while preserving enough trace-level detail and repair opportunities.\\
\addlinespace
40, iter. 10 &
The final batch has 0/40 successes after earlier train batches reached as high as 57.5\%; the saved update records a 40/40 \texttt{ConfigError}, where 16 previously passing tasks and 24 failing tasks all become exceptions. &
An iteration-9 middleware rewrite left the YAML config using \texttt{max\_repetitions}, while the progress-monitor middleware expects \texttt{max\_iterations}; iteration 10 aligns the parameter name. &
Broad shared failures become easy to detect, but the update is dominated by a global contract mismatch.\\
\addlinespace
80, iter. 5 &
Train-batch success stays nontrivial across B1--B5 (42.5\%, 45.0\%, 42.5\%, 40.0\%, 43.8\%), yet final validation and ID regress. The update finds that 25/29 failed HLE tasks use \texttt{write\_file} answer submission that bypasses an answer-verification guard. It also records pass-to-fail regressions including \texttt{fix-git}, \texttt{hle\_\_66ecb2...}, \texttt{hle\_\_67206a...}, \texttt{largest-eigenval}, scheduler, and \texttt{password-recovery}. &
The guard previously checked shell-command answer writes; the update extends detection to \texttt{write\_file} and \texttt{replace} and scans written content for substantive verification. The unrelated regressions reflect different mechanisms, including final-state mismatch, over-engineering, unmet optimization constraints, and hallucinated recovery from partial data. &
The most salient global pattern can dominate diverse task-specific failures.\\
\addlinespace
80 continuation &
Validation from the batch-80 final snapshot shows typed-message runtime failures, and the first continuation 80-task train job has 26 errored trials. &
The continuation starts from an unstable runtime state rather than a clean harness. &
Large batches can leave unresolved execution-path instability in the selected snapshot.\\
\bottomrule
\end{tabular}
\caption{Representative update artifacts from the batch-size sweep. The cases explain the non-monotonic result in Table~\ref{tab:batch-size-case-summary}.}
\label{tab:batch-size-case-evidence}
\end{table*}

The cases in Table~\ref{tab:batch-size-case-evidence} support a more specific reading than a monotonic scaling law.
Batch 10 receives the largest update budget per train task, about 0.31M tokens, so its failure is not simply a lack of analysis budget.
The iteration-16 update targets a real rabbit-hole pattern, but the evidence is narrow: only one non-exception task failure drives the substantive harness change.
By iteration 40, a single constructor/configuration mismatch prevents all ten tasks from starting.
The negative result is therefore better explained as update-stream instability: frequent small-batch edits can learn local failure patterns while also accumulating runtime and configuration contract risk.

Batch 20 is the only setting with large positive validation and ID gains, but it is not risk-free.
Its intermediate $E_{16}$ replay collapse shows that a runtime-path failure can affect most tasks.
The important difference is that the batch-20 schedule still leaves enough evidence and enough later update opportunities to identify and repair the shared message-content contract failure.
The final update restores the middleware guidance path, current-batch success recovers from 1/20 to 10/20, and replay errors fall from 66 to 5.
Thus, batch 20 is strong in this run because the evidence is diverse enough to reveal cross-task runtime problems, but still small enough for trace-level inspection.

Batch 40 illustrates the intermediate regime.
It has enough evidence to expose broad shared failures, and it finishes with a small positive validation and ID gain.
However, its per-task update budget is only about 40\% of batch 20's, so task-specific analysis is thinner.
The final 40/40 \texttt{ConfigError} is easy to identify as a shared contract mismatch, but the update is largely spent restoring the runtime path rather than learning more heterogeneous task improvements.

Batch 80 is the clearest evidence-overload case.
The train batches themselves do not collapse immediately, but each update must summarize 80 heterogeneous trajectories under roughly the same 3--4M-token update budget.
Iteration 5 finds an important HLE-wide blind spot in answer-file verification, yet the same batch also contains unrelated regressions on implementation, optimization, and recovery tasks.
With only five update opportunities, a broad middleware change around the most salient pattern has little room for later correction.
The continuation run confirms that the selected final snapshot carries an unstable runtime state rather than merely a low score.

\begin{table*}[t]
\centering
\scriptsize
\setlength{\tabcolsep}{3.4pt}
\begin{tabular}{@{}p{0.10\textwidth}p{0.43\textwidth}p{0.39\textwidth}@{}}
\toprule
\textbf{Setting} & \textbf{Concrete evidence} & \textbf{Mechanism}\\
\midrule
Batch 10 &
Iteration 16 has one substantive task failure and adds rabbit-hole detection around \texttt{regex-chess}; iteration 40 produces 10/10 \texttt{ConfigError} from a \texttt{max\_interventions} constructor mismatch. &
Evidence is narrow and updates are frequent, so local rules can accumulate configuration and runtime-contract risk.\\
\addlinespace
Batch 20 &
Iteration 20 repairs message-content construction for middleware guidance; validation, ID, and OOD all improve in the selected run. &
Evidence diversity and per-trace analysis depth are balanced; the schedule can detect cross-task runtime failures and still recover.\\
\addlinespace
Batch 40 &
Iteration 10 detects a 40/40 \texttt{ConfigError} from a \texttt{max\_repetitions}/\texttt{max\_iterations} mismatch; final gains are small but positive. &
Shared failures are visible, but per-task analysis depth is lower than batch 20, limiting heterogeneous task improvement.\\
\addlinespace
Batch 80 &
Iteration 5 finds that 25/29 failed HLE tasks bypass answer verification through file tools, while multiple unrelated pass-to-fail regressions remain; final and continuation diagnostics show an unstable runtime state. &
Evidence overload makes the most salient global pattern visible, but long-tail failures are diluted and broad middleware edits have fewer later repair opportunities.\\
\bottomrule
\end{tabular}
\caption{Mechanistic summary of the batch-size case study. AHE update quality depends on evidence density, per-trace analysis depth, update frequency, and opportunities to repair earlier harness edits.}
\label{tab:batch-size-mechanism-summary}
\end{table*}

Overall, the result should not be read as a simple claim that larger or smaller batches are intrinsically better.
AHE update is not a mini-batch gradient step; it is an LLM-driven harness-editing process with an approximately fixed token and attention budget per update.
Batch size changes the amount and heterogeneity of evidence that must be integrated under that budget.
Small batches expose too little evidence and require many edits; very large batches expose many failures but dilute per-task analysis and encourage broad changes around the most visible pattern.
Batch 20 is best in this sweep because it jointly provides diverse failure evidence, enough trace-level detail for diagnosis, and enough update opportunities to repair runtime or configuration contract mistakes introduced earlier.

\subsection{Source Diversity Case Study}
\label{app:source-diversity-case-study}

The source-diversity experiment compares two AHE training streams with the same train size, batch size, and number of epochs.
Both runs expose 80 train tasks, use batch size 20, and perform 20 updates.
The difference is the evidence source.
The mixed-source run uses Terminal-Bench plus HLE, so its update evidence contains implementation, tool-use, environment, file-operation, long-context, and HLE reasoning failures.
The HLE-only run mostly exposes knowledge, math, physics, and answer-verification failures.
Table~\ref{tab:source-diversity-case-summary} summarizes the outcome.

\begin{table*}[t]
\centering
\small
\setlength{\tabcolsep}{4.2pt}
\begin{tabular}{@{}p{0.25\textwidth}rrrp{0.28\textwidth}@{}}
\toprule
\textbf{Source setting} & \textbf{Final validation} & \textbf{Final ID} & \textbf{Final OOD} & \textbf{Selected checkpoint}\\
\midrule
Terminal-Bench + HLE & 57.1 & 49.1 & 28.8 & Final $A_T$\\
HLE only & 0.0 & 0.0 & 0.0 & $E_{12}$: ID 47.3, OOD 25.0\\
\bottomrule
\end{tabular}
\caption{Source-diversity case-study summary. The HLE-only final snapshot collapses, but its intermediate $E_{12}$ snapshot is useful; we therefore analyze both the intermediate gains and the final failure.}
\label{tab:source-diversity-case-summary}
\end{table*}

We do not read the HLE-only result as ``no learning.''
The HLE-only run produces a useful intermediate snapshot: validation rises from 40.0\% at $E_0$ to 42.9\% at $E_{12}$, the shared ID view reaches 47.3\%, and OOD reaches 25.0\%.
The final snapshot then collapses to 0.0\% on validation, ID, and OOD.
Our interpretation is therefore more specific: HLE-only evidence can produce locally useful HLE-specific harness improvements, but the later update stream pushes the harness toward a fragile verification and message-injection path that the final snapshot does not preserve.

\begin{table*}[t]
\centering
\scriptsize
\setlength{\tabcolsep}{3.4pt}
\begin{tabular}{@{}p{0.30\textwidth}p{0.32\textwidth}p{0.30\textwidth}@{}}
\toprule
\textbf{Evidence} & \textbf{Harness update} & \textbf{Mechanism}\\
\midrule
At iteration 11, the current 20-task HLE batch has 14 failures and 6 successes, with no execution exceptions. The evolve summary assigns 10/14 failures to circular verification: the agent writes verification code, but the code reproduces the same flawed derivation. &
The self-verification middleware records whether \texttt{run\_code} is called after rejection and injects a stronger rejection when the agent retries without new code. &
The update no longer asks only for verification; it requires the post-rejection verification path to change.\\
\addlinespace
After middleware rejection, the agent often rereads the problem and resubmits the same answer. &
The rejection message is strengthened to require an independent method and red-team testing. &
We move the failure response from prompt advice into middleware-level interaction control.\\
\addlinespace
The HLE failures require more specific checking strategies than a generic ``verify again'' instruction. &
The update changes the system prompt, the \texttt{run\_code} tool description, and long-term memory. &
The HLE-specific failure pattern becomes visible in the later harness state.\\
\bottomrule
\end{tabular}
\caption{HLE-only iteration-11 evidence. The run quickly turns HLE failures into self-verification mechanisms rather than tool or environment repairs.}
\label{tab:source-diversity-hle-iter11}
\end{table*}

Table~\ref{tab:source-diversity-hle-iter11} explains why the HLE-only run can improve before it collapses.
The update artifacts show AHE learning to block premature HLE answers, circular verification, and self-confirming reasoning.
This is a real harness improvement, but it is concentrated in the answer-verification loop.
In other words, HLE-only evidence gives us dense signal about one subsystem and much less signal about the rest of the runtime interaction path.

\begin{table*}[t]
\centering
\scriptsize
\setlength{\tabcolsep}{3.4pt}
\begin{tabular}{@{}p{0.28\textwidth}p{0.34\textwidth}p{0.30\textwidth}@{}}
\toprule
\textbf{Observed failure} & \textbf{Update response} & \textbf{Why it explains the intermediate gain}\\
\midrule
The agent's verification code and written reasoning share the same wrong assumption. &
After rejection, the middleware requires new code execution and emphasizes an independent method. &
This reduces cases where an incorrect answer is ``verified'' by a script that encodes the same mistake.\\
\addlinespace
An LLM critic fails for more than five iterations; all three implementation attempts return empty content, and fallback templates are too generic. &
Iteration 18 removes the LLM critic and replaces it with a rule-based domain analyzer keyed to reliability, physics, quantum, math, geometry, multiple choice, and related domains. &
The feedback becomes more stable and more domain-specific than the empty critic path.\\
\addlinespace
After rejection, the agent tends to continue along the original answer path. &
The update adds a \texttt{before\_model} hook that injects a structured, domain-specific verification reminder before the next model response. &
The harness can intervene before the next generation rather than only after another premature submission.\\
\bottomrule
\end{tabular}
\caption{HLE-only intermediate-gain evidence. The useful $E_{12}$ checkpoint is consistent with saved updates that intensify the self-verification loop.}
\label{tab:source-diversity-hle-intermediate}
\end{table*}

The $E_{12}$ checkpoint is therefore not just curve noise.
Its gains match the artifacts in Table~\ref{tab:source-diversity-hle-intermediate}: the run has been optimizing rejection handling, domain-specific checking, and the timing of verification reminders.
Because the source stream is HLE-only, the resulting harness improvement is also narrow.
It improves how the agent handles HLE-style answer verification, but it gives little direct evidence about file-state constraints, tool failures, long-running commands, environment setup, or implementation artifacts.

\begin{table*}[t]
\centering
\scriptsize
\setlength{\tabcolsep}{3.4pt}
\begin{tabular}{@{}p{0.26\textwidth}p{0.34\textwidth}p{0.32\textwidth}@{}}
\toprule
\textbf{Final-collapse evidence} & \textbf{Runtime cause} & \textbf{Interpretation}\\
\midrule
From iteration 19 to 20, pass rate falls from 7/20 to 0/20, and all 20 tasks become \texttt{RuntimeError}. &
The error concentrates on the NexAU token counter: it now accepts only \texttt{Sequence[Message]}, while legacy dictionary messages are no longer supported. &
The final collapse is not a gradual loss of HLE reasoning ability; the tasks are blocked by the runtime message contract.\\
\addlinespace
Iteration 18 introduced a \texttt{before\_model} hook for self-verification reminders. &
The hook inserts plain dictionaries such as \texttt{\{"role": "user", "content": text\}} into the message list. &
The same middleware path that produced stronger HLE verification also creates the message-injection failure.\\
\addlinespace
Iteration 20 changes the injected message to \texttt{Message.user(text)}. &
The repair targets the message type rather than the HLE answer strategy itself. &
The failure is located in the verification middleware's runtime interface.\\
\bottomrule
\end{tabular}
\caption{HLE-only final-collapse evidence. The final snapshot fails because self-verification middleware violates the typed-message runtime contract.}
\label{tab:source-diversity-hle-collapse}
\end{table*}

The HLE-only final snapshot shows the cost of this narrow optimization path.
To repair circular verification, AHE repeatedly edits self-verification middleware.
That middleware directly injects messages before the model is called, so a message-construction error affects every HLE task before task solving can proceed.
Once that path violates the runtime contract, validation, ID, and OOD all go to zero.
This is why we avoid saying only that single-source training ``overfits.''
The more precise failure mode is that a single reasoning-heavy source concentrates updates on one verification subsystem, and the final runtime failure occurs exactly in that subsystem's message-injection path.

The mixed-source run also passes through a bad intermediate state, so source diversity is not a guarantee against harmful updates.
The difference is the evidence available for recovery.
Terminal-Bench exposes tool calls, file editing, environment setup, long-running commands, artifact cleanup, and tool-error recovery.
HLE exposes reasoning, web evidence, verification sufficiency, and multiple-choice checking.
In the mixed-source artifacts, early updates add file tools, web tools, session-lifecycle tools, and context compaction; later updates add HLE verification enforcement, artifact cleanup, search-failure fallback, contradictory-evidence handling, and message-content contract alignment.
When a message-contract failure appears, both Terminal-Bench and HLE tasks are affected, which helps us see it as a shared execution-path failure rather than as an HLE reasoning failure.
After the final message-content repair, the current batch recovers from 1/20 to 10/20 and train-replay errors fall from 66 to 5.
The final mixed-source snapshot then preserves gains on validation, ID, and OOD.

\begin{table*}[t]
\centering
\scriptsize
\setlength{\tabcolsep}{3.4pt}
\begin{tabular}{@{}p{0.22\textwidth}p{0.38\textwidth}p{0.32\textwidth}@{}}
\toprule
\textbf{Setting} & \textbf{Concrete evidence} & \textbf{Mechanism}\\
\midrule
HLE-only intermediate $E_{12}$ &
Iteration 11 repairs circular verification; iteration 18 replaces a failing LLM critic with a rule-based domain analyzer. &
Single-source HLE evidence can produce useful HLE-specific verification improvements.\\
\addlinespace
HLE-only final &
Iteration 20 records 20/20 \texttt{RuntimeError} from plain-dictionary message insertion into a typed-message runtime. &
Later updates concentrate on self-verification middleware and eventually break its message contract.\\
\addlinespace
Mixed-source final &
Saved updates cover tools, web/session/context management, HLE verification, artifact cleanup, fallback search, contradictory evidence, and message-content alignment; final validation, ID, and OOD are all positive. &
The mixed source exposes both reasoning failures and execution failures, so harness changes cover more of the runtime interaction path and the run has evidence to recover shared runtime failures.\\
\addlinespace
HLE-only updates2 diagnostic &
The diagnostic run has no complete final evaluation, but its validation moves from 37.1 at $E_0$ to 40.0 at $E_4$/$E_8$, drops to 8.6 at $E_{12}$, and recovers to 40.0 at $E_{16}$. &
More update opportunities can allow self-repair, but they do not remove the instability caused by a narrow HLE verification evidence source.\\
\bottomrule
\end{tabular}
\caption{Mechanistic summary of source diversity. We find that the source determines which harness subsystem receives evidence, and that subsystem becomes both the likely improvement target and the likely failure point.}
\label{tab:source-diversity-mechanism-summary}
\end{table*}

Our conclusion is therefore not that mixed data is universally better.
For AHE, the update target is the harness, and the type of harness-failure evidence strongly determines where updates are applied.
HLE-only data gives dense reasoning and verification failures, so AHE learns to intensify self-verification; this can produce useful intermediate gains, but it also concentrates risk in the same message-injection path.
Mixed-source data gives a more distributed set of failures, including tools, environments, files, long-running execution, context pressure, and HLE reasoning.
That diversity does not prevent harmful updates, but in this run it helps the final harness recover from an intermediate collapse and retain gains across validation, ID, and OOD.

\subsection{Cross-Model Case Study}
\label{app:cross-model-case-study}

The cross-model setting lets us inspect how the same AHE update process changes when the rollout backend changes.
We do not infer the mechanism from the final score alone.
Instead, we read the update artifacts from each run: at every update, the evolve agent receives the current batch trajectories, groups failure patterns, and edits prompts, memory, or middleware.
These artifacts show that the three rollout backends expose different failure surfaces, and AHE consequently modifies different harness subsystems.
Table~\ref{tab:cross-model-case-summary} summarizes the evidence before we discuss the concrete updates.

\begin{table*}[t]
\centering
\scriptsize
\setlength{\tabcolsep}{3.5pt}
\begin{tabular}{@{}p{0.13\textwidth}p{0.31\textwidth}p{0.25\textwidth}p{0.23\textwidth}@{}}
\toprule
\textbf{Rollout backend} & \textbf{Recurring trajectory evidence} & \textbf{Main updated components} & \textbf{Update tendency}\\
\midrule
DeepSeek &
Buggy verification, missing verification, artifact residue, repeated heredocs, sleep chaining, and middleware message-contract errors. &
\texttt{HLEVerificationMiddleware}, \texttt{ToolErrorRecoveryMiddleware}, \texttt{VerificationReminderMiddleware}, prompts, and memory. &
Repair both reasoning verification and the runtime interaction path.\\
\addlinespace
GLM &
HLE tasks produce text-only reasoning without reading instructions or writing answer files; research-heavy tasks spend many turns researching without creating the requested output. &
\texttt{TaskTypeOptimizerMiddleware}, task-type workflows, and research-loop reminders. &
Move the agent from explanation or research into tool action and output production.\\
\addlinespace
GPT-5.4 &
Answer-only HLE tasks are under-classified; artifacts violate self-contained constraints; validation is attempted but not passing; single-file tasks leave build byproducts. &
\texttt{ExecutionGuardMiddleware}, artifact scanning, and validation-state tracking. &
Check whether the final artifact satisfies the task contract and whether validation actually passed.\\
\bottomrule
\end{tabular}
\caption{Cross-model case-study summary. Different rollout backends expose different trajectory evidence, and AHE edits the harness subsystem that matches that evidence.}
\label{tab:cross-model-case-summary}
\end{table*}

\paragraph{DeepSeek.}
In the DeepSeek run, updates cover a broad runtime interaction path.
Iteration 9 observes that several HLE failures are not caused by the absence of verification.
The agent writes verification code, but the code encodes the same wrong assumptions as the written reasoning, so the previous middleware accepts a self-confirming script as evidence.
The same update also records an implementation-side artifact issue: \texttt{polyglot-c-py} leaves an extra compiled binary in the solution directory.
The resulting patch strengthens \texttt{HLEVerificationMiddleware}: a single script and a numerical output no longer count as sufficient evidence, and the harness asks for web research, independent computation, an eval/check script, or multiple separate computations.
The update also records artifact cleanup in the system prompt and \texttt{LongTermMEMORY}.

Later updates show that the DeepSeek trajectories expose more than answer-level errors.
Iteration 14 finds that \texttt{ToolErrorRecoveryMiddleware} is not firing because it accesses a nonexistent \texttt{context\_overflow\_imminent} field.
The same batch contains execution-pattern failures: \texttt{path-tracing} repeatedly uses large heredocs for complex code, and \texttt{rstan-to-pystan} enters long chained sleeps.
The patch repairs the middleware field access and adds runtime detection for heredoc-heavy coding and sleep chaining.
Iteration 20 then repairs a shared message-contract failure: earlier middleware created \texttt{Message(..., content="string")}, but the NexAU schema requires block-structured content, so the patch changes the injected messages to \texttt{content=[TextBlock(...)]}.
Together, these updates explain why the DeepSeek-evolved harness contains transferable execution constraints: verification, tool recovery, artifact cleanup, long-running execution control, and message injection all enter the editable harness state.

\paragraph{GLM.}
The GLM run exposes a different failure surface.
In iteration 16, three HLE tasks produce only text reasoning: the agent does not read \texttt{instruction.md} and does not write an answer file.
In \texttt{protein-assembly}, the agent spends 28 messages on PDB/fpbase research but never creates the required \texttt{gblock.txt}.
These trajectories show a workflow-entry failure: the agent remains in explanation or research mode instead of entering the task's required execution path.

The update therefore does not primarily extend the HLE verifier.
It restores and strengthens task-type workflows in the prompt and memory, covering HLE, implementation, image, research-heavy, and package-installation tasks.
It also edits \texttt{TaskTypeOptimizerMiddleware}: HLE tasks can be detected from the initial user message, workflow guidance is injected on the first turn, text-only HLE responses trigger a stronger read/write reminder, and research-heavy non-HLE tasks receive a progress reminder once the agent has used tools for several turns without producing the requested output.
This is a concrete workflow-control update, not a generic instruction to reason better.
The GLM trajectories push AHE toward action forcing and output production because those are the failures visible in the batch.

The next update also shows the risk of this path.
Iteration 17 reports that the newly edited middleware injects legacy dictionary messages, while the runtime expects typed \texttt{Message} objects.
The repair converts dictionary injections in \texttt{task\_type\_optimizer.py} and \texttt{invalid\_tool\_call.py} into \texttt{Message(role=Role.FRAMEWORK, content=[TextBlock(...)])} and adds helper functions that read either typed messages or dictionaries.
The same workflow reminder that helps move GLM rollouts into action depends on message injection; when that contract is wrong, the harness improvement becomes an execution-path error.

\paragraph{GPT-5.4.}
The GPT-5.4 run concentrates on artifact constraints and validation sufficiency.
In iteration 12, two answer-only HLE failures are under-classified: prompts such as direction-choice or counting questions do not match the previous strict \texttt{Question:/Answer:/Confidence:} pattern, so the stronger quantitative workflow is not reliably injected.
The same update finds that \texttt{path-tracing} writes \texttt{/app/image.c} that delegates to \texttt{/app/orig} through \texttt{execl}, violating the self-contained task requirement.
It also finds that \texttt{regex-chess} attempts \texttt{/app/check.py} late in the rollout, times out, and never records a passing behavior-level validation.
The resulting \texttt{execution\_guard.py} patch expands answer-task markers, tracks forbidden-read paths and external-helper signals such as \texttt{/app/orig}, \texttt{execl}, \texttt{system}, and \texttt{subprocess}, and distinguishes attempted validation from passing validation.

Iteration 17 continues the same pattern.
For \texttt{polyglot-c-py}, the raw trace shows validation with \texttt{gcc /app/polyglot/main.py.c -o /app/polyglot/cmain}, which leaves a compiled byproduct in a directory where the task asks for a single submitted file.
For \texttt{install-windows-3-11}, the observed HTTP response redirects to \texttt{http://127.0.0.1/...}; local checks may appear successful, but a remote verifier cannot follow a loopback URL.
The update again modifies \texttt{execution\_guard.py}, adding a single-file validation-byproduct guard and a loopback-redirect guard.
Thus, the GPT-5.4-evolved harness mainly learns to inspect whether the produced artifact and validation trace satisfy the task contract, rather than simply asking the model to deliberate longer.

\begin{table*}[t]
\centering
\scriptsize
\setlength{\tabcolsep}{3.4pt}
\begin{tabular}{@{}p{0.12\textwidth}p{0.34\textwidth}p{0.28\textwidth}p{0.18\textwidth}@{}}
\toprule
\textbf{Run} & \textbf{Concrete trajectory evidence} & \textbf{Harness update} & \textbf{Mechanism}\\
\midrule
DeepSeek iter. 9 &
HLE verification scripts reproduce the same wrong assumptions; \texttt{polyglot-c-py} leaves an extra binary. &
Stricter HLE verification and artifact-cleanup rules. &
Verification and final-state constraints become executable harness behavior.\\
\addlinespace
DeepSeek iter. 14--20 &
Tool recovery middleware is broken; heredoc and sleep-chain patterns waste turns; message content violates the typed-message schema. &
Tool-error repair, heredoc/sleep-chain detection, and \texttt{TextBlock}-based message construction. &
Shared runtime interaction paths are repaired.\\
\addlinespace
GLM iter. 16 &
HLE tasks produce only text reasoning; \texttt{protein-assembly} researches for 28 messages without writing \texttt{gblock.txt}. &
Task-type optimizer, first-turn workflow guidance, text-only HLE action reminders, and research-loop progress checks. &
The harness pushes the agent from explanation or research into actions and output files.\\
\addlinespace
GLM iter. 17 &
New workflow middleware injects legacy dictionary messages and triggers runtime errors. &
Dictionary injections are converted into typed \texttt{Message} objects with \texttt{TextBlock} content. &
The workflow-control path is restored by repairing the message contract.\\
\addlinespace
GPT-5.4 iter. 12 &
Answer-only HLE tasks are missed; \texttt{path-tracing} delegates to \texttt{/app/orig}; \texttt{regex-chess} attempts but never passes validation. &
Expanded answer-task detection, forbidden-helper scanning, and attempted-vs-passing validation tracking. &
The harness checks artifact constraints and validation sufficiency.\\
\addlinespace
GPT-5.4 iter. 17 &
\texttt{polyglot-c-py} leaves a compile byproduct; \texttt{install-windows-3-11} redirects remote users to loopback. &
Single-file byproduct cleanup guard and loopback-redirect guard. &
Runtime-observed false positives are turned into narrow execution guards.\\
\bottomrule
\end{tabular}
\caption{Representative cross-model update artifacts. The examples show how each rollout backend exposes different evidence and therefore induces different AHE harness edits.}
\label{tab:cross-model-update-artifacts}
\end{table*}

These artifacts explain the asymmetric ID and OOD results in Table~\ref{tab:cross-model-transfer-full}.
AHE's gains first depend on whether the harness edits match the failure surface of the evaluation trajectories.
The DeepSeek-evolved harness repairs verification, tool recovery, artifact cleanup, and message contracts, so it obtains a same-backend ID gain and can transfer some general execution constraints to other backends.
The GLM-evolved harness mainly addresses text-only reasoning and research without output; it is most useful when the evaluation trajectory lacks action progress, and less useful when the dominant failures are artifact-contract violations.
The GPT-5.4-evolved harness focuses on artifact inspection and validation sufficiency; it improves same-backend ID, but its guards are tied to concrete tool outputs and artifact patterns, so their benefit is less stable when the backend or source distribution changes.

The same evidence also explains why ID and OOD gains diverge.
An update can be well aligned with the failures observed during source training but miss the shifted failures in another rollout backend or target domain.
When evaluation trajectories still contain similar patterns, such as faulty verification, tool-recovery failure, missing output files, or validation false positives, the corresponding harness change can transfer.
When the main failures shift to patterns not exposed during training, the gain shrinks or is offset by extra reminders and guards.
Thus, the cross-model results support a mechanism-level conclusion: same-backend ID gains are easier because the evaluation trajectories reuse failure surfaces observed during training, whereas cross-backend and OOD evaluations require the learned harness edits to survive both model-behavior shift and task-distribution shift.

\FloatBarrier

\section{Additional Results}
\label{app:additional-results}

This section provides the supplementary plots referenced by the experiment section.
The main text reports compact epoch-level curves and summary tables; the figures below expose source-group breakdowns, batch-index views, and replay-based fix/forgetting diagnostics.
These plots use the same saved task-result records and aggregation conventions as Section~\ref{sec:experiments}.

\begin{center}
\centering
\includegraphics[width=\linewidth]{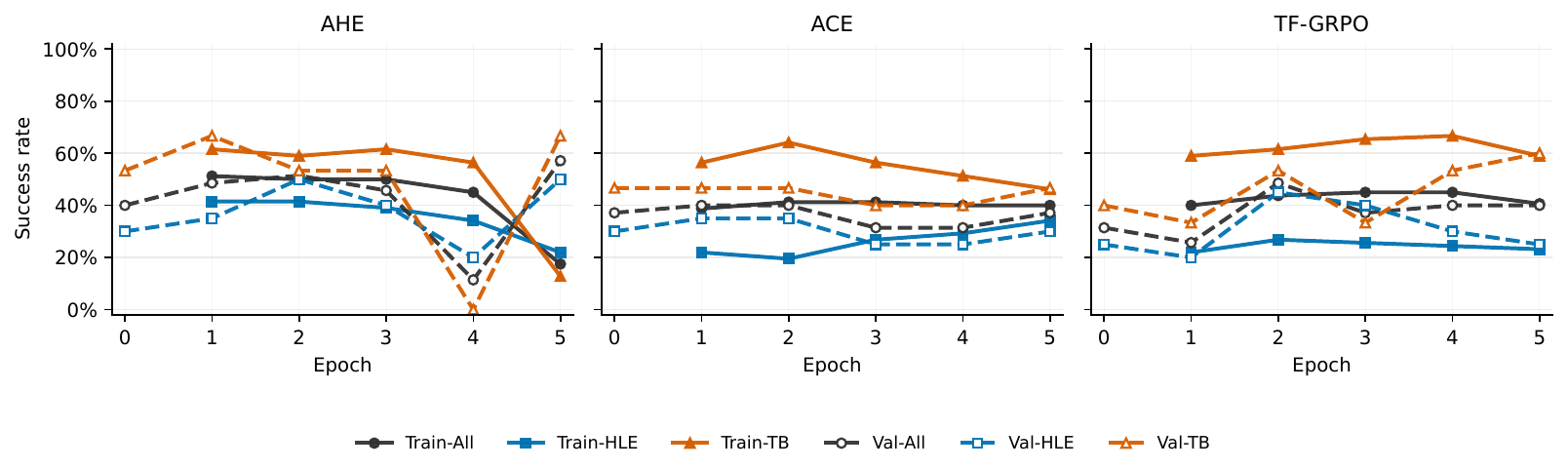}
\captionof{figure}{Baseline success-rate breakdown by source group. Each panel reports one method; colors distinguish all tasks, HLE tasks, and Terminal-Bench tasks, while line style distinguishes train and validation curves.}
\label{fig:baseline-results-breakdown-curves}
\end{center}

\begin{center}
\centering
\includegraphics[width=\linewidth]{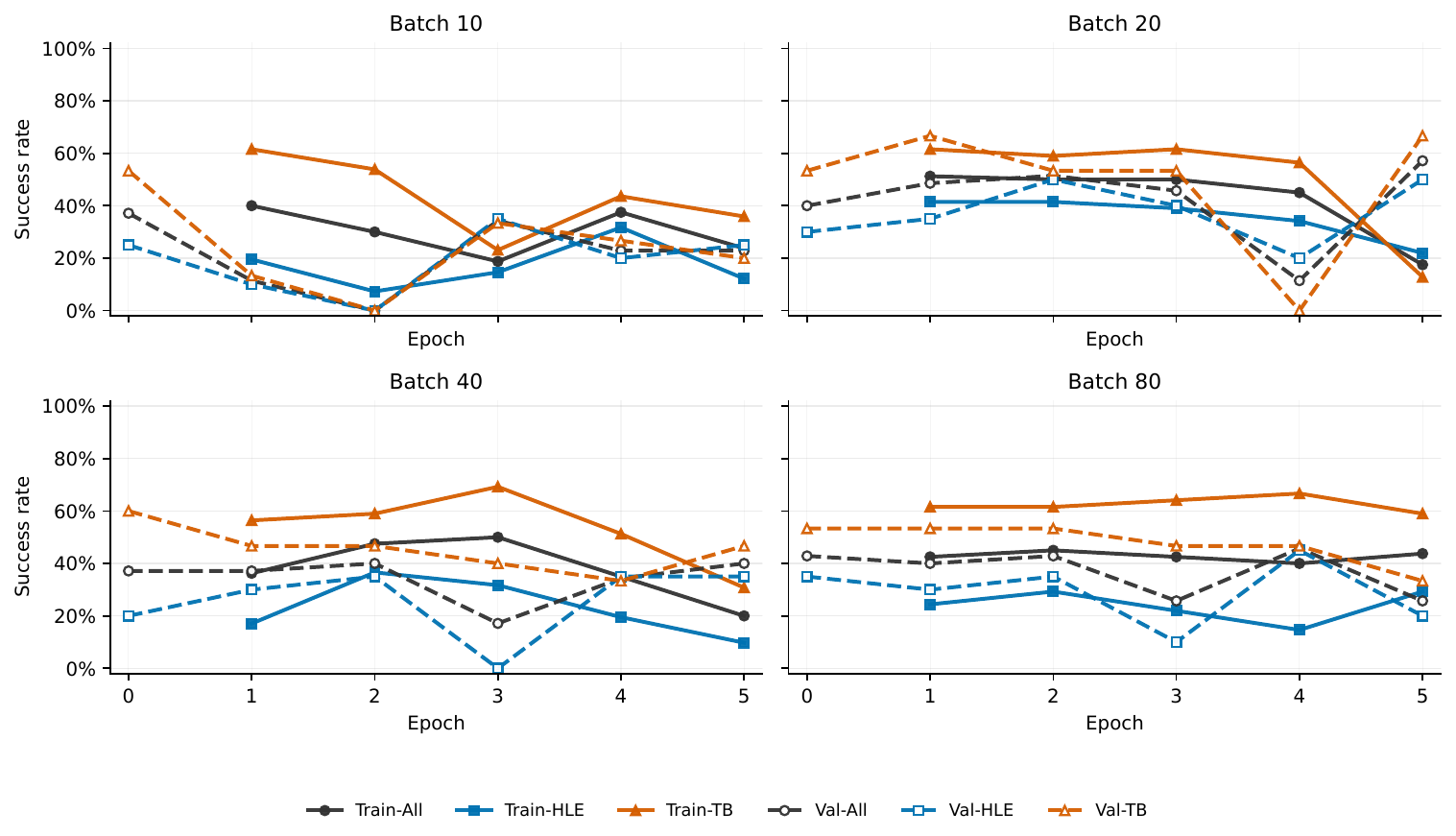}
\captionof{figure}{AHE batch-size success-rate breakdown by source group. Each panel reports one batch size; colors distinguish all tasks, HLE tasks, and Terminal-Bench tasks, while line style distinguishes train and validation curves.}
\label{fig:batch-size-breakdown-curves}
\end{center}

\begin{center}
\centering
\includegraphics[width=\linewidth]{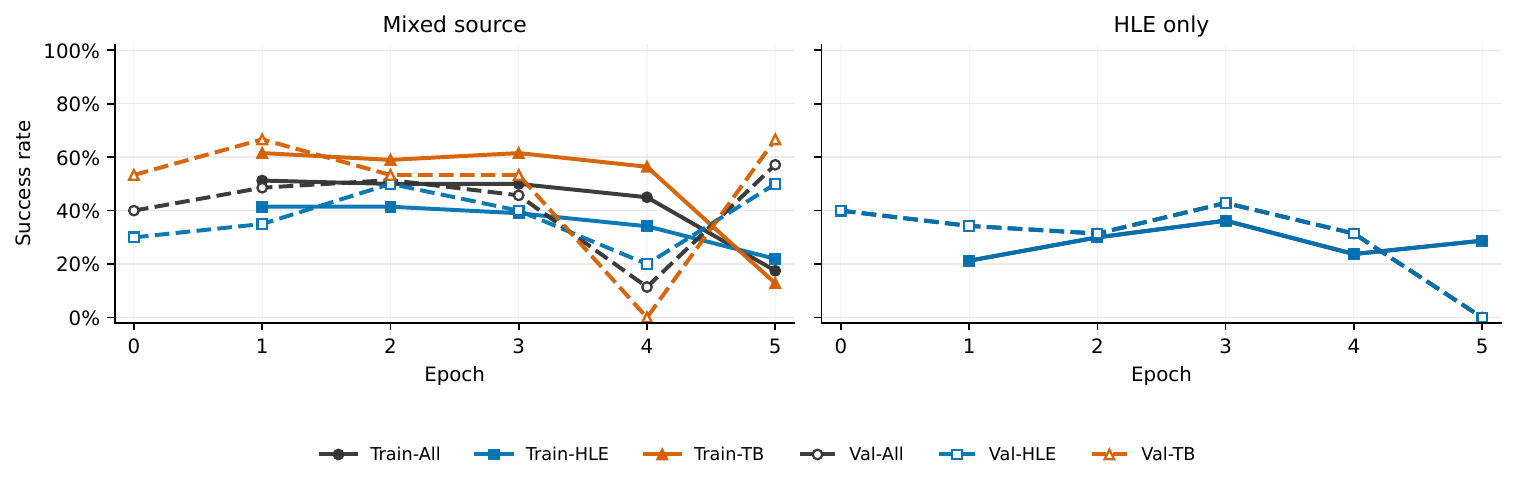}
\captionof{figure}{AHE source-diversity success-rate breakdown by source group. Each panel reports one source setting; colors distinguish all tasks, HLE tasks, and Terminal-Bench tasks, while line style distinguishes train and validation curves.}
\label{fig:source-diversity-breakdown-curves}
\end{center}

\begin{center}
\centering
\includegraphics[width=\linewidth]{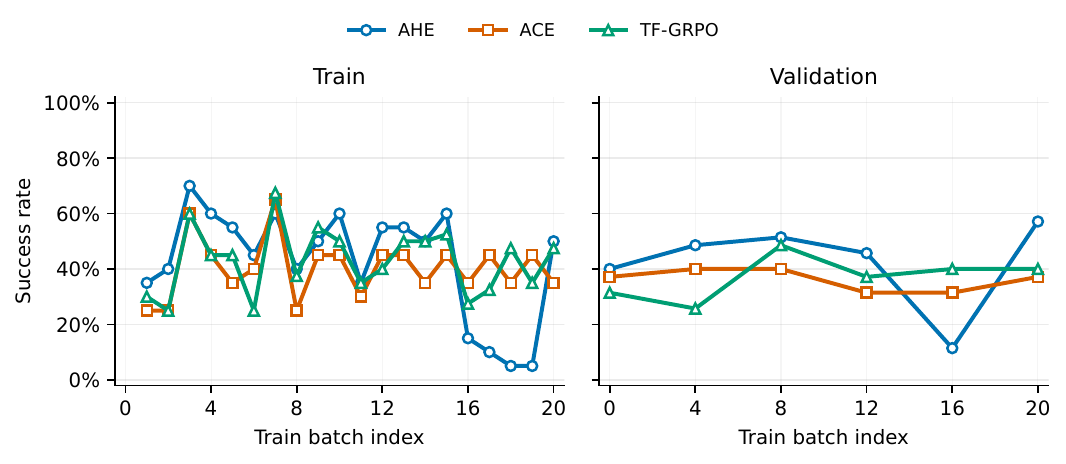}
\captionof{figure}{Baseline learning curves by train batch index. Validation points are plotted at the corresponding epoch-end batch index.}
\label{fig:baseline-results-batch-index-curves}
\end{center}

\begin{center}
\centering
\includegraphics[width=\linewidth]{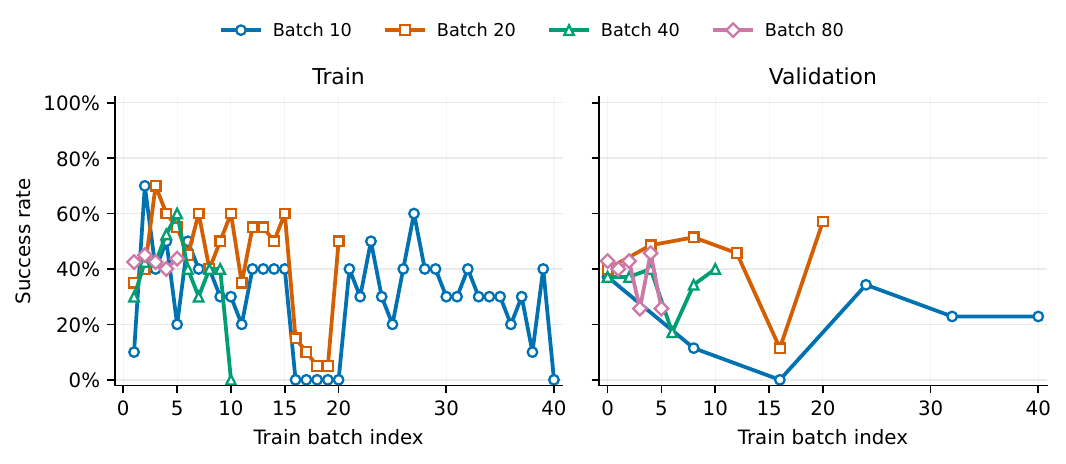}
\captionof{figure}{AHE batch-size learning curves by train batch index. Different batch sizes have different numbers of update points; validation points are plotted at the corresponding epoch-end batch index.}
\label{fig:batch-size-batch-index-curves}
\end{center}

\begin{center}
\centering
\includegraphics[width=\linewidth]{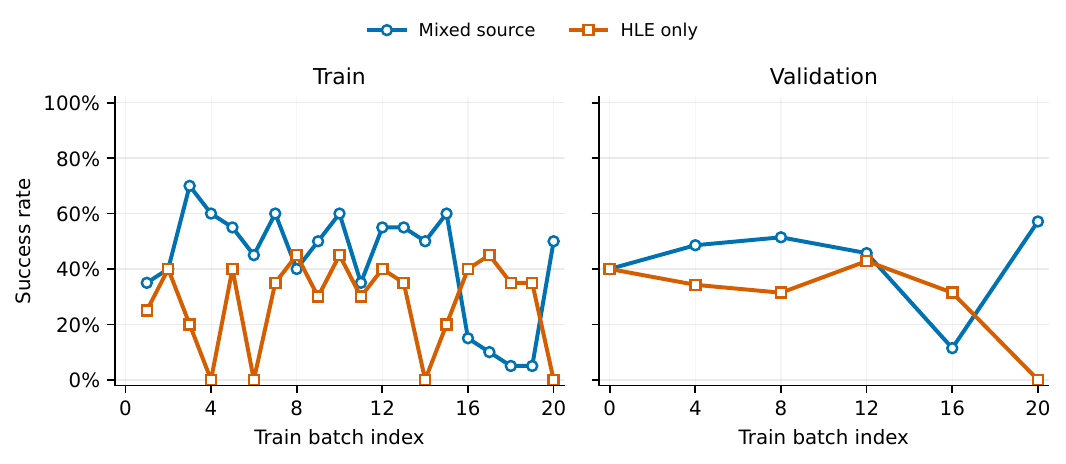}
\captionof{figure}{AHE source-diversity learning curves by train batch index. Validation points are plotted at the corresponding epoch-end batch index.}
\label{fig:source-diversity-batch-index-curves}
\end{center}

\subsection{Cross-Model Continuation Results}
\label{app:cross-model-continuation}

The cross-model appendix compares AHE training runs using the same Terminal-Bench + HLE source setting and batch-20 schedule, with DeepSeek-V4-Flash, GLM-5.1, and GPT-5.4 as the training backend.
The GPT-5.4 row uses the continuation run from epoch 3 through epoch 5.
Table~\ref{tab:cross-model-continuation} reports the training-run validation and cost fields; Figure~\ref{fig:cross-model-transfer-heatmap} summarizes the final ID and OOD gains, and Table~\ref{tab:cross-model-transfer-full} reports the underlying success rates.

\begin{center}
\centering
\includegraphics[width=\linewidth]{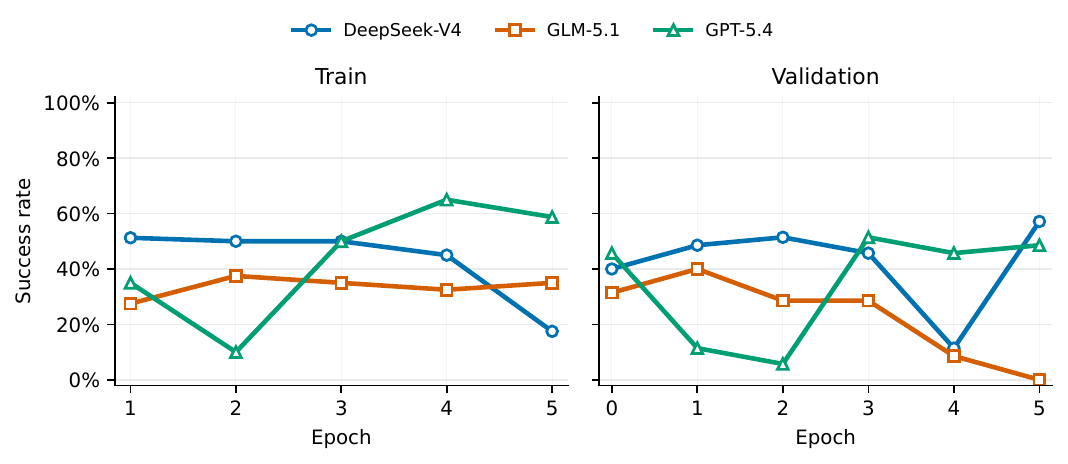}
\captionof{figure}{AHE cross-model learning curves. The left panel reports epoch-averaged train success rate, and the right panel reports epoch-end validation success rate.}
\label{fig:cross-model-curves}
\end{center}

\begin{center}
\centering
\includegraphics[width=\linewidth]{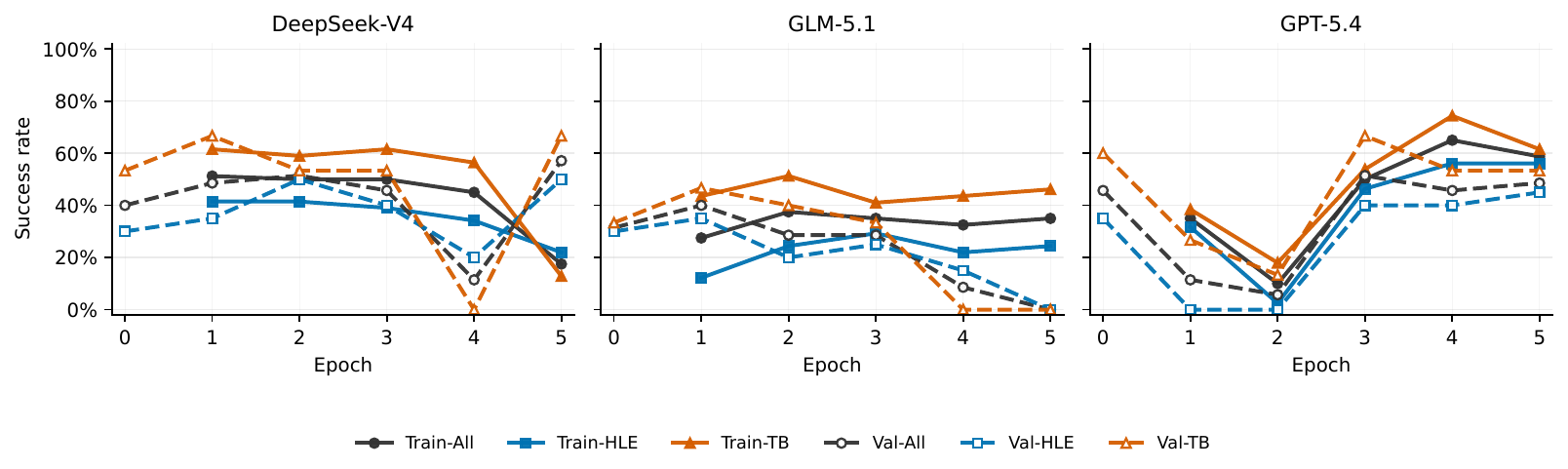}
\captionof{figure}{AHE cross-model success-rate breakdown by source group. Each panel reports one training backend; colors distinguish all tasks, HLE tasks, and Terminal-Bench tasks, while line style distinguishes train and validation curves.}
\label{fig:cross-model-breakdown-curves}
\end{center}

\begin{center}
\centering
\tiny
\setlength{\tabcolsep}{2.0pt}
\resizebox{\linewidth}{!}{%
\begin{tabular}{@{}lrrrrrrlll@{}}
\toprule
\textbf{Training backend} & \textbf{$V_0$} & \textbf{$V_T$} &
\textbf{UVG} & \textbf{$V_{\max}$} & \textbf{Best epoch} &
\textbf{Rollout tok./task} & \textbf{Update tok./update} &
\textbf{Updates} & \textbf{Status}\\
\midrule
DeepSeek-V4-Flash & 40.0 & 57.1 & +17.1 & 57.1 & 5 & 1.46M & 3.91M & 20/20 changed & complete\\
GLM-5.1 & 31.4 & 0.0 & -31.4 & 40.0 & 1 & 0.22M & 5.55M & 19/20 changed & final validation collapse\\
GPT-5.4 & 45.7 & 48.6 & +2.9 & 51.4 & 3 & 0.19M & 3.62M & 17/20 changed & continuation complete\\
\bottomrule
\end{tabular}
}
\captionof{table}{AHE cross-model continuation summary. Success-rate columns are percentages, and UVG is the final validation gain in percentage points. $V_{\max}$ is the best epoch-end validation score observed during training. Token costs are normalized as rollout tokens per evaluated task/trial and recorded update tokens per SEAGym update call. Final ID and OOD results are reported separately in Table~\ref{tab:cross-model-transfer-full}.}
\label{tab:cross-model-continuation}
\end{center}

\begin{center}
\centering
\tiny
\setlength{\tabcolsep}{1.8pt}
\resizebox{\linewidth}{!}{%
\begin{tabular}{@{}llrrrrrr@{}}
\toprule
\textbf{Evolved harness} & \textbf{Rollout model} &
\textbf{ID$_0$} & \textbf{ID$_\star$} & \textbf{IDG} &
\textbf{OOD$_0$} & \textbf{OOD$_\star$} & \textbf{OODG}\\
\midrule
DeepSeek-evolved (epoch 5) & DeepSeek & 40.0 & 49.1 & +9.1 & 22.5 & 28.8 & +6.3\\
DeepSeek-evolved (epoch 5) & GLM & 45.5 & 52.7 & +7.3 & 23.8 & 23.8 & 0.0\\
DeepSeek-evolved (epoch 5) & GPT-5.4 & 47.3 & 43.6 & -3.6 & 38.8 & 30.0 & -8.8\\
\addlinespace
GLM-evolved (epoch 1) & DeepSeek & 40.0 & 47.3 & +7.3 & 22.5 & 17.5 & -5.0\\
GLM-evolved (epoch 1) & GLM & 45.5 & 49.1 & +3.6 & 23.8 & 27.5 & +3.8\\
GLM-evolved (epoch 1) & GPT-5.4 & 47.3 & 47.3 & 0.0 & 38.8 & 33.8 & -5.0\\
\addlinespace
GPT-5.4-evolved (epoch 3) & DeepSeek & 40.0 & 38.2 & -1.8 & 22.5 & 26.3 & +3.8\\
GPT-5.4-evolved (epoch 3) & GLM & 45.5 & 38.2 & -7.3 & 23.8 & 22.5 & -1.3\\
GPT-5.4-evolved (epoch 3) & GPT-5.4 & 47.3 & 52.7 & +5.5 & 38.8 & 31.3 & -7.5\\
\bottomrule
\end{tabular}
}
\captionof{table}{Full cross-model ID and OOD results. Success rates are percentages. Gains compare the selected evolved snapshot to the same rollout model's $A_0$ result on the same evaluation set.}
\label{tab:cross-model-transfer-full}
\end{center}

\subsection{Train Replay Fix and Forgetting Metrics}
\label{app:fix-forgetting-metrics}

The train replay diagnostics in Figure~\ref{fig:fix-forgetting-summary} use two complementary fix/forget definitions.
Let $T$ be the replay task set, $S_e \subseteq T$ be the tasks solved after epoch $e$, and $S_0$ be the tasks solved by $A_0$ before training.

The pairwise delta metrics count task churn between adjacent snapshots:
\[
\Delta\mathrm{Fix}_e = |S_e \setminus S_{e-1}|,\qquad
\Delta\mathrm{Forget}_e = |S_{e-1} \setminus S_e|.
\]
These quantities are defined for $e>0$; the initial agent is plotted with zero delta.
We report these as task counts rather than rates because their purpose is local process diagnosis: how many individual tasks were newly fixed or broken by the latest update interval.
Using changing success/failure denominators would mix task churn with denominator drift and make the recovery after epoch 4 harder to interpret.

The $A_0$-reference metrics compare every snapshot to the fixed initial agent:
\[
\mathrm{Fix}^{A_0}_e = \frac{|S_e \setminus S_0|}{|T \setminus S_0|},\qquad
\mathrm{Forget}^{A_0}_e = \frac{|S_0 \setminus S_e|}{|S_0|}.
\]
These rates answer a different question: relative to the initial harness, what fraction of initially failed tasks has been fixed, and what fraction of initially solved tasks has been lost?
The denominators are fixed across epochs, so the curves are directly comparable over time.
All replay diagnostics are computed offline by the benchmark and are not fed back to the evolving agent.
Appendix~\ref{app:training-forgetting-case-study} analyzes the corresponding saved snapshots, task trajectories, and update artifacts.

\begin{figure*}[t]
\centering
\includegraphics[width=0.98\textwidth]{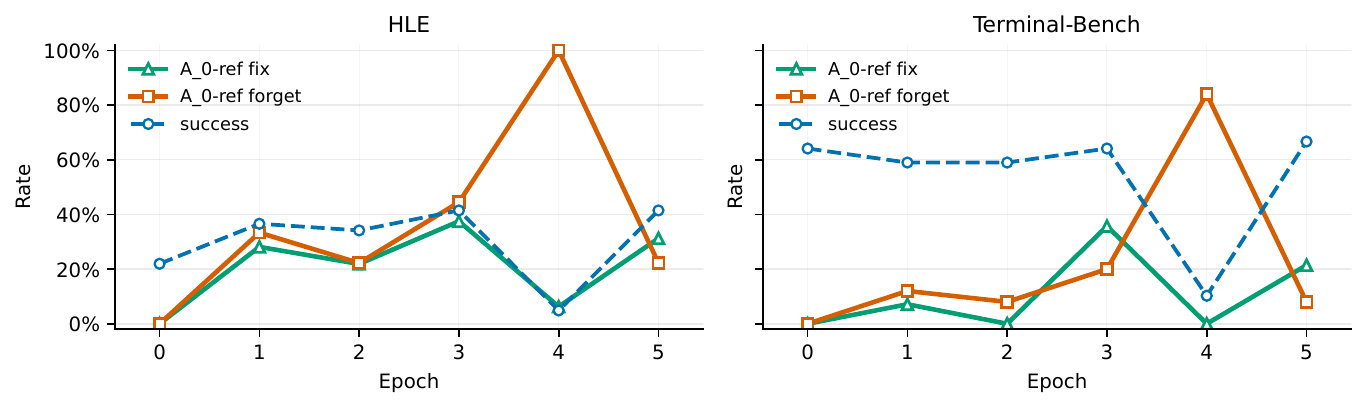}
\caption{Source-group train replay diagnostics for AHE. Each panel reports success rate and $A_0$-reference fix/forget rates for HLE or Terminal-Bench tasks using source-specific fixed denominators.}
\label{fig:fix-forgetting-by-source}
\end{figure*}

\FloatBarrier

\section{Integration Details}
\label{app:integration-details}
\enlargethispage{3\baselineskip}

\subsection{Task Index and Visibility}

\textsc{SEAGym} uses a lightweight task index rather than copying benchmark task definitions.
Each indexed task stores a stable id, source benchmark reference, task attributes, scoring metadata, and visibility metadata.
Executable instructions, environments, verifiers, and raw artifacts remain in the underlying benchmark backend whenever possible.

The agent-visible task view excludes private evaluation metadata, such as reference outputs, private assertions, split membership, and held-out view labels.
This separation prevents evaluation metadata from becoming update evidence.

\subsection{Rollout and Update Interfaces}

The method interface has two roles:
\begin{itemize}
  \item \textbf{rollout adapter}: runs a batch of tasks under the current harness state and returns trajectories, verifier rewards, public errors, and observed cost;
  \item \textbf{update adapter}: consumes the trajectory batch, applies the method's native update rule, and saves updated harness state and update summaries.
\end{itemize}
This split is needed because many benchmark runners instantiate task agents per trial, while self-evolution state must persist outside individual task executions.
It also allows different update mechanisms to share the same task schedule and assessment protocol.

\subsection{Harbor Backend}

The implementation uses Harbor as the benchmark execution substrate.
Harbor runs tasks, environments, verifiers, parallel jobs, and trial artifacts.
\textsc{SEAGym} adds the outer self-evolution schedule, snapshot records, held-out view orchestration, metric inputs, and normalized reports.
The integration does not require modifying Harbor source code or redefining Harbor's adapter standard.

\section{Metric Details}
\label{app:metric-details}

\begin{table}[t]
\centering
\tiny
\renewcommand{\arraystretch}{1.08}
\begin{tabular}{@{}p{0.11\linewidth}p{0.34\linewidth}p{0.43\linewidth}@{}}
\toprule
\textbf{Metric} & \textbf{Definition} & \textbf{Interpretation}\\
\midrule
Perf & $\frac{1}{|D|}\sum_{x\in D} r(A,x)$ & normalized verified task performance\\
SR & $\frac{1}{|D|}\sum_{x\in D}\mathbb{I}[r(A,x)=1]$ & success rate when task scores are binary\\
UVG$^{\text{prev}}$ & $\text{Perf}(E_i,V)-\text{Perf}(E_{i-1},V)$ & local update-validation gain\\
UVG$^{\text{base}}$ & $\text{Perf}(E_i,V)-\text{Perf}(E_0,V)$ & cumulative update-validation gain\\
IDG & $\text{Perf}(A_T,D_I)-\text{Perf}(A_0,D_I)$ & final held-out in-domain gain\\
OODG & $\text{Perf}(A_T,D_O)-\text{Perf}(A_0,D_O)$ & final OOD transfer gain\\
FR & $\max(0,\text{Perf}(A_0,D_R)-\text{Perf}(A_T,D_R))$ & replay forgetting or regression\\
Cost & tokens, tool calls, time, or dollars & efficiency and overhead of evolution\\
\bottomrule
\end{tabular}
\caption{Summary of \textsc{SEAGym} metrics. Primary scores come from verified task outcomes; update-validation, ID, OOD, and replay metrics explain the trajectory of self-evolution.}
\label{tab:metrics}
\end{table}

In these experiments, task scores are binary, so we report the
aggregate verified outcome as a success rate. The saved evaluation summaries
also include a stricter execution-success flag, which additionally requires
that no execution exception was recorded. We use the verified task score for
paper metrics and reserve execution exceptions, timeouts, provider failures,
and middleware errors for process diagnostics.

\subsection{Saved Records}

Metrics are computed from saved run records rather than live objects.
The run stores evaluation-point summaries, task-level normalized results, verifier outputs, cost records, update summaries, snapshot references, and backend job references.
This lets users recompute metrics or change aggregation rules without rerunning task environments.

\subsection{Aggregation}

Main tables use domain-level macro averages by default so that domains with more tasks do not dominate the score.
Micro averages and task-level breakdowns can be reported in appendix tables for diagnostics.

\end{document}